\documentclass{article} %
\usepackage{iclr2024_conference,times}

\usepackage{amsmath,amsfonts,bm}

\def\eqref#1{equation~\ref{#1}}

\def\1{\bm{1}}

\DeclareMathAlphabet{\mathsfit}{\encodingdefault}{\sfdefault}{m}{sl}
\SetMathAlphabet{\mathsfit}{bold}{\encodingdefault}{\sfdefault}{bx}{n}

\usepackage{xcolor,soul}
\usepackage{hyperref}
\usepackage{url}
\usepackage{booktabs}
\usepackage{multirow}
\usepackage{multicol}
\usepackage{subcaption}
\usepackage{graphicx}
\usepackage{amssymb}
\usepackage{float}

\newcommand{\method}[1]{{\texttt{FireAct}}}
\newcommand{\react}[1]{{\texttt{ReAct}}}
\newcommand{\reflexion}[1]{{\texttt{Reflexion}}}
\newcommand{\cott}[1]{{CoT}}

\title{\method{}: Toward Language Agent Fine-tuning}

\author{Baian Chen\thanks{Equal contribution. Code, data, and models are available at \url{https://fireact-agent.github.io}.} \\
System2 Research\\
\And
Chang Shu$^*$ \\
University of Cambridge\\
\And
Ehsan Shareghi\\
Monash University \\
\AND
Nigel Collier\\
University of Cambridge \\
\And
Karthik Narasimhan\\
PLI, Princeton University \\
\And
Shunyu Yao \\
PLI, Princeton University \\
}

\iclrfinalcopy %
\begin{document}

\maketitle

\begin{abstract}
Recent efforts have augmented language models (LMs) with external tools or environments, leading to the development of \emph{language agents} that can reason and act. However, most of these agents rely on few-shot prompting techniques with off-the-shelf LMs. 
In this paper, we investigate and argue for  the overlooked direction of fine-tuning LMs to obtain language agents.
Using a setup of question answering (QA) with a Google search API,  
we explore a variety of base LMs, prompting methods, fine-tuning data, and QA tasks, 
and find language agents are consistently improved after fine-tuning their backbone LMs.
For example, fine-tuning Llama2-7B with 500 agent trajectories generated by GPT-4 leads to a 77\% HotpotQA performance increase.
Furthermore, we propose \method{}, a novel approach to fine-tuning LMs with trajectories from multiple tasks and prompting methods, and show having more diverse fine-tuning data can further improve agents. 
Along with other findings regarding 
scaling effects, robustness, generalization, efficiency and cost,
our work establishes comprehensive benefits of fine-tuning LMs for agents, and provides an initial set of experimental designs, insights, as well as open questions toward language agent fine-tuning.

\end{abstract}

\section{Introduction}
\label{sec:introduction}
Recent work has explored grounding language models~\citep[LMs;][]{brown2020language, chowdhery2022palm, touvron2023llama} to interact with external tools or environments, leading to a new class of \emph{language agents}~\citep{nakano2021webgpt, yao2022react, park2023generative} that could obtain new knowledge from environmental feedback, make sequential decisions via language reasoning, and improve task solving using self-reflection~\citep{shinn2023reflexion, wang2023voyager}. 
Beyond research,  industrial developments  such as  ChatGPT Plugins~\citep{plugins} 
have indicated the great potential of language agents for real-world applications.

So far, most language agents prompt off-the-shelf LMs for convenience and flexibility. However, existing LMs  were not developed for agentic usecases (e.g., generating actions or self-evaluations), for which few-shot prompting only offers limited learning support. As a result, most LMs have poor performance and robustness when used for agents, and some advanced agents~\citep{yao2023tree, wang2023voyager} can only be supported by GPT-4~\citep{openai2023gpt4}, resulting in high costs and latencies, along with issues like controllability and reproducibility. 

\begin{figure}[t]
    \centering
    \includegraphics[width=\linewidth]{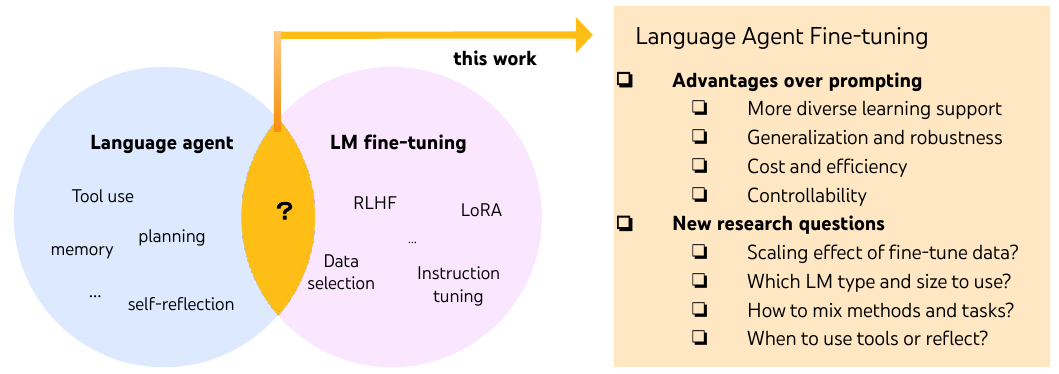}
    \caption{While language agents and language model fine-tuning are both popular topics, their intersection is understudied. This work takes an initial step to show multiple advantages of fine-tuning LMs for agentic uses, and opens up various new questions toward language agent fine-tuning.}
    \label{fig:teaser}
\end{figure}

Fine-tuning is an appropriate solution for these issues: it has been shown that fine-tuned smaller LMs could outperform prompted larger LMs for specific reasoning~\citep{zelikman2022star, huang2022large} and acting~\citep{yao2022react} needs, while enjoying reduced inference time and expense. But the study of LM fine-tuning for agents has been very limited, despite the large amount of studies around language agents and LM fine-tuning respectively (Figure~\ref{fig:teaser}). 
Only a few prior works have fine-tuned LMs for web navigation~\citep{nakano2021webgpt, yao2022webshop} or API tool use~\citep{schick2023toolformer,patil2023gorilla, qin2023toolllm}, with preliminary scaling analysis specific to a type of models~\citep{yao2022react,schick2023toolformer,nakano2021webgpt}.

In this work, we take an initial step toward a more systematic study of language agent fine-tuning.
We propose \method{}, a novel way to fine-tune LMs with agent trajectories generated from multiple tasks and prompting methods, and unified in the \react{}~\citep{yao2022react} format (Figure~\ref{fig:method}).
We implement \method{} using open-domain question answering (QA) tasks with access to a Google search API, and GPT-4~\citep{openai2023gpt4} for fine-tuning data generation.
By thoroughly investigating a variety of base LMs~\citep{openai2023finetune, touvron2023llama, Rozire2023CodeLO}, prompting methods~\citep{yao2022react, wei2022chain, shinn2023reflexion}, fine-tuning data, and tasks~\citep{yang-etal-2018-hotpotqa, press2022measuring, DBLP:conf/iclr/HendrycksBBZMSS21, DBLP:journals/tacl/GevaKSKRB21}, our experiments illustrate various advantages of fine-tuning and the importance of fine-tuning data diversity. For example, while few-shot \react{} prompting GPT-3.5 on HotpotQA achieves an exact match (EM) score of 31.4, fine-tuning with 500 \react{} trajectories improves the EM to 39.2 (25\% increase), and fine-tuning with a mix of \react{} and \cott{} trajectories further improves the EM to 41.0 (31\% increase). Furthermore, fine-tuning reduces inference time by 4x, and improves performances by 64\% in face of distracting tool outputs. Such benefits can be even more visible for smaller open-source LMs where few-shot prompting performs poorly, e.g., fine-tuning Llama2-7B~\citep{touvron2023llama} leads to a 77\% EM increase on HotpotQA.

Besides showcasing these benefits, our experiments also explore complex interactions among various factors of fine-tuning and provide actionable insights for practitioners. As for the base LM, we find GPT-3.5 significantly outperforms other open-source LMs when fine-tuning with less than 500 samples, but the gap can be gradually caught up by scaling to more fine-tuning samples. As for the prompting methods to generate fine-tuning data, we find different LMs benefit from different mix ratios, and present trajectory statistics and oracle analyses for further understanding. As for the tasks to generate fine-tuning data, our preliminary results show that adding a task might not improve downstream performances on significantly different tasks, but also does not hurt performances. This suggests the potential for massive multi-task fine-tuning to obtain a single LM as the agent backbone for various applications. Along with various other findings, discussions, and the release of \method{} code, data, and  model checkpoints, we hope our work ignites and inspires future efforts toward more capable and useful fine-tuned language agents.

\section{Related Work}
\label{sec:related}
\textbf{Language agents}. Language agents~\citep{weng2023prompt,wang2023survey} represent an emerging kind of AI systems that use language models (LMs) to interact with the world. While earliest language agents simply used LMs to generate action commands~\citep{nakano2021webgpt,huang2022language, ahn2022can, schick2023toolformer}, learning direct observation-action mappings from few-shot demonstrations is challenging when the domain is complex or involves long-horizon activities.
ReAct~\citep{yao2022react} proposed to use LMs to generating both reasoning traces~\citep{wei2022chain, nye2021work, DBLP:conf/nips/KojimaGRMI22} and actions, so that reasoning can flexibly guide, track, and adjust acting, leading to substantial improvements over act-only methods. 
Follow up work has applied LM-based reasoning for more purposes in agent design, such as reflection~\citep{shinn2023reflexion, park2023generative}, planning~\citep{yao2023tree, dagan2023dynamic, liu2023llm+}, program synthesis~\citep{liang2023code, wang2023voyager}, etc. The forms of external grounding have also diversified, ranging from digital games~\citep{huang2022language, wang2023voyager}, APIs~\citep[``tools'';][]{schick2023toolformer, patil2023gorilla, qin2023toolllm}, webpages~\citep{yao2022webshop, deng2023mind2web, zhou2023webarena}, to physical~\citep{DBLP:journals/corr/abs-2309-01918, DBLP:journals/corr/abs-2306-17582, DBLP:conf/icml/DriessXSLCIWTVY23}, human~\citep{zhang2020dialogpt}, and multi-agent~\citep{park2023generative} interactions. We refer readers to~\citet{xi2023rise} for an empirical survey and~\citet{sumers2023cognitive} for a systematic theoretical framework of language agents. Notably, most existing language agents prompted off-the-shelf LMs.

\textbf{Language model fine-tuning.} Adapting pre-trained LMs to downstream tasks is another active field of study~\citep{DBLP:journals/corr/abs-2308-10792}, including various instruction-based fine-tuning datasets~\citep{DBLP:conf/acl/MishraKBH22, DBLP:conf/iclr/SanhWRBSACSRDBX22, DBLP:journals/corr/abs-2304-07327,DBLP:conf/acl/WangKMLSKH23,DBLP:conf/acl/HonovichSLS23,longpre2023flan}, models~\citep{alpaca, vicuna2023, DBLP:journals/corr/abs-2304-12244, DBLP:conf/acl/MuennighoffWSRB23, DBLP:conf/nips/Ouyang0JAWMZASR22}, parameter-efficient fine-tuning methods~\citep{DBLP:conf/iclr/HuSWALWWC22, DBLP:journals/natmi/DingQYWYSHCCCYZWLZCLTLS23, DBLP:journals/corr/abs-2306-09782, DBLP:journals/corr/abs-2305-14314, DBLP:conf/acl/IvisonBWHP23}, and data selection principles ~\citep{DBLP:journals/corr/abs-2305-11206, gunasekar2023textbooks}.
Additionally, there are various studies on fine-tuning specific types of LMs, such as coding LMs~\citep{Li2023StarCoderMT, DBLP:journals/corr/abs-2306-08568, DBLP:journals/corr/abs-2308-12950}, multi-modal LMs~\citep{DBLP:journals/corr/abs-2306-17107, DBLP:journals/corr/abs-2305-04790, DBLP:journals/corr/abs-2305-06500, DBLP:journals/corr/abs-2306-02858, DBLP:conf/cvpr/BrooksHE23, DBLP:journals/corr/abs-2305-16355}, and retrieval-augmented LMs~\citep{DBLP:conf/icml/GuuLTPC20, DBLP:journals/corr/abs-2304-08085}. However, fine-tuning LMs for language agents that reason and act has been limited.

\textbf{Language agent fine-tuning.} Despite the vast interests in language agents and  fine-tuning, their intersection has received limited attention, with only some initial study about how performances scale with the model size for a particular model family~\citep{nakano2021webgpt, schick2023toolformer, yao2022react}, how to incorporate more tools via retrieval~\citep{patil2023gorilla,qin2023toolllm}, and some task-specific ablations~\citep{yao2022webshop, le2022coderl}. This paper takes on a more systematic investigation, proposing and answering new questions toward language agent fine-tuning.

\section{\method{}: Fine-tuning LMs with Diverse \react{} Trajectories}
\label{sec:setups}

\begin{figure}[t]
    \centering
    \includegraphics[width=\linewidth]{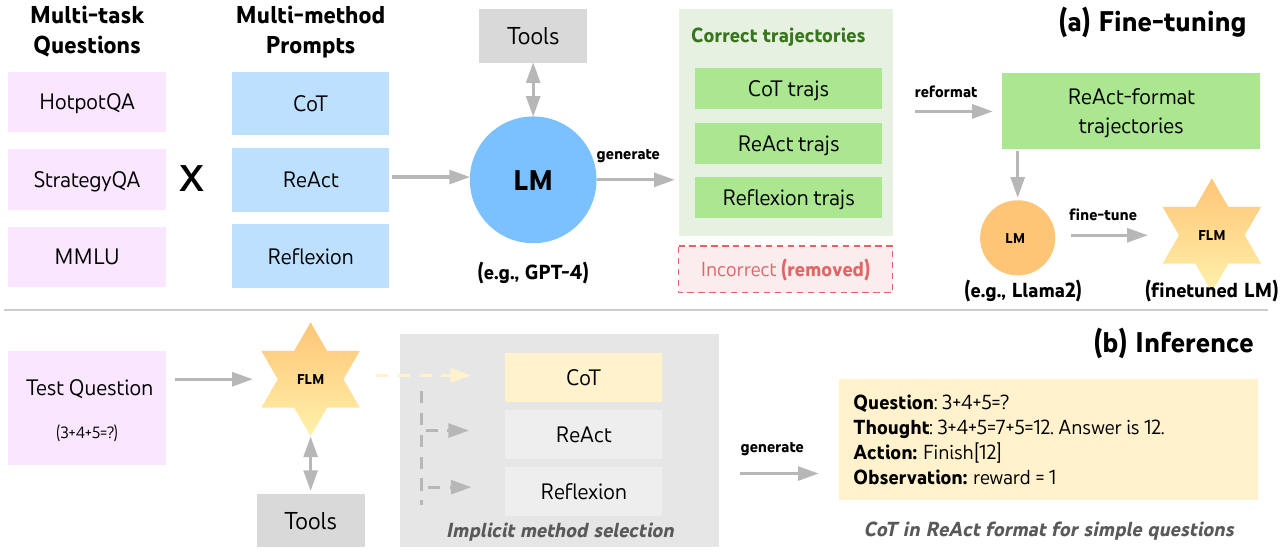}
    \caption{Illustration of \method{}. (a) \textbf{During fine-tuning}, a large LM (e.g., GPT-4) generates task-solving trajectories based on questions from different datasets and prompts from different methods. The successful trajectories are then converted into the \react{} format to fine-tune a smaller LM. (b) \textbf{During inference}, the fine-tuned LM could operate without few-shot prompting, and could implicitly select an prompting method to complete a \react{} trajectory with flexible lengths, adapting to different question complexities. 
    For example, a simple question could be solved using only one thought-action-observation round, without using tools.
    }
    \label{fig:method}
\end{figure}

Our work is largely based on \react{}~\citep{yao2022react}, a popular approach to language agents. A \react{} task-solving trajectory (Figure~\ref{fig:traj}) consists of multiple thought-action-observation rounds, where an LM generates free-form ``thoughts'' for versatile purposes (e.g., extract information from observations, propose and adjust action plans, track task progress), and structured ``actions'' to interact with   environments (tools) and receive ``observation'' feedback. 
\react{} outperforms reasoning or acting only baselines, as reasoning can guide acting, and acting can support reasoning with new information. 
The \react{} format has thus been a basis of many follow-up language agents, such as Reflexion~\citep{shinn2023reflexion}, SwiftSage~\citep{lin2023swiftsage}, and AutoGPT~\citep{autogpt}.

Also shown in~\citep{yao2022react} was a preliminary PaLM~\citep{chowdhery2022palm} fine-tuning experiment on HotpotQA~\citep{yang-etal-2018-hotpotqa}, where a fine-tuned PaLM-62B outperforms a prompted PaLM-540B. But it remains unknown if such a finding generalizes to other types of LMs, prompting methods, or tasks. Follow-up studies on language agent fine-tuning have been sparse (see Section~\ref{sec:related}).

Thus we propose \method{}, a novel fine-tuning approach to language agents. As shown in Figure~\ref{fig:method}(a), \method{} also leverages few-shot prompting of a strong LM to generate diverse \react{} trajectories  to fine-tune a smaller LM (i.e., distillation~\citep{hinton2015distilling}). But different from~\citet{yao2022react}, \method{} explicitly promotes data diversity by mixing multiple training tasks and prompting methods. Here we consider two other  methods compatible with the \react{} format:
\begin{itemize}
\item \textbf{Chain of Thought (\cott{})}~\citep{wei2022chain} generates intermediate reasoning to bridge the question-answer gap. Each \cott{} trajectory can be turned into a simple one-round \react{} trajectory, with ``thought'' being the intermediate reasoning and ``action'' being returning the answer. \cott{} is useful for simple questions without tool needs (Figure~\ref{fig:method}(b)).%
\item \textbf{Reflexion}~\citep{shinn2023reflexion} mostly follows the \react{} trajectory, but incorporates extra feedback and self-reflections. In this work, we simply prompt for reflections at the 6th and 10th \react{} round, so that long \react{} trajectories could pivot the strategy for solving the current task (e.g., ``film search has not been helpful yet, I should search directors now'').
\end{itemize}

During inference (Figure~\ref{fig:method}(b)), a \method{} agent alleviates the need for few-shot prompting, which makes inference more efficient and convenient. It could also implicitly select the suitable method adaptive to the task complexity, and show stronger generalization and robustness than prompting as a result of a wider and more diverse learning support.

\section{Experimental Setup}

\textbf{Tasks}. Following prior work~\citep{wei2022chain,yao2022react, shinn2023reflexion}, we train and test on well-established question answering (QA) tasks, which enjoy abundant and high-quality training data plus easy and faithful evaluation (answer exact match). We use four datasets:
\begin{itemize}
    \item 
    \textbf{HotpotQA}~\citep{yang-etal-2018-hotpotqa} is a QA dataset challenging multi-step reasoning and knowledge retrieval.  The answer is usually a short entity or yes/no. We use 2,000 random training questions for fine-tuning data curation, and 500 random dev questions for evaluation.
    \item 
    \textbf{Bamboogle}~\citep{press2022measuring} is a test set of 125 multi-hop questions with similar formats as HotpotQA, but carefully crafted to avoid direct solving with Google search.
    \item 
    \textbf{StrategyQA}~\citep{DBLP:journals/tacl/GevaKSKRB21} is a yes/no QA dataset requiring implicit reasoning steps. 
    \item 
    \textbf{MMLU}~\citep{DBLP:conf/iclr/HendrycksBBZMSS21} covers 57 multi-choice QA tasks in various domains such as elementary mathematics, history, and computer science. 
\end{itemize}

\textbf{Tool}. Following~\citet{press2022measuring}, we use SerpAPI\footnote{\url{https://serpapi.com}.} to build a Google search tool that returns the first existent item from ``answer box'', ``answer snippet'', ``highlight words'', or ``first result snippet'', which ensures the response is short and relevant. We find such a simple tool sufficient for basic QA needs across tasks, and increases our fine-tuned models' ease of use and generality.

\textbf{LMs}. We investigate three families of LMs: 
\begin{itemize}
\item \textbf{OpenAI GPT}. We prompt GPT-4~\citep{openai2023gpt4} to generate all fine-tuning data, and use GPT-3.5 for fine-tuning~\citep{openai2023finetune} as well as prompting. We used both models in ChatCompletion mode from July to Sept 2023. 
\item \textbf{Llama-2}~\citep{DBLP:journals/corr/abs-2307-09288} with 7B and 13B parameters in ``chat'' mode.
\item \textbf{CodeLlama}~\citep{DBLP:journals/corr/abs-2308-12950} with 7B, 13B, and 34B parameters in ``instruct'' mode, which help further understand model size scaling and the importance of code fine-tuning for agentic tasks.
\end{itemize}

\textbf{Fine-tuning methods}. We use Low-Rank Adaptation (LoRA)~\citep{DBLP:conf/iclr/HuSWALWWC22} for most fine-tuning experiments, but also use full-model fine-tuning for some comparison.

Given the various factors underlying language agent fine-tuning,
We split experiments into three parts with increasing complexities:
\begin{itemize}
    \item Fine-tuning using a single prompting method on a single task  (Section~\ref{sec:ss});
    \item Fine-tuning using multiple methods on a single task (Section~\ref{sec:multi-method});
    \item Fine-tuning using multiple methods on multiple tasks (Section~\ref{sec:multi-task}). 
\end{itemize}

\section{Single-task, Single-method Fine-tuning}
\label{sec:ss}

In this section, we focus on fine-tuning with data from a single task (HotpotQA) and a single prompting method (\react{}). Using such a simple and controlled setup, we confirm various benefits of fine-tuning over prompting (performance, efficiency, robustness, generalization), and study effects of different LMs, data sizes, and fine-tuning methods.
By default, we use 500 successful few-shot prompting trajectories generated by GPT-4 for training and a random subset of 500 HotpotQA dev questions for evaluation. 
Other experimental details can be found in the Appendix~\ref{app:exp_detail}. 
\subsection{Performance and efficiency}

\begin{table}[ht]
\centering
\begin{minipage}[t]{0.3\linewidth}
\centering
\caption{Prompting results.}
\begin{tabular}{lcc}
\toprule
                    & Prompt        & EM            \\ \hline
\multirow{3}{*}{GPT-4}         & IO                     & 37.2          \\
                               & \cott{}                    & \textbf{45.0} \\
                               & \react{}                  & 42.0          \\ \hline
\multirow{3}{*}{GPT-3.5} & IO                     & 22.4          \\
                               & \cott{}                    & 28.0          \\
                               & \react{}                  & \textbf{31.4} \\ 

\bottomrule
\end{tabular}
\label{tab:hotpot_baseline}
\end{minipage}%
\begin{minipage}[t]{0.7\linewidth}
\centering
\caption{Prompting vs.\,fine-tuning, with absolute/relative increases.}
\begin{tabular}{lcccc}
\toprule
            & \react{}        & \method{} &  abs./rel. diff\\ \hline
Llama-2-7B& 14.8                     & 26.2        & +11.4 / \textbf{77\%}\\
Llama-2-13B& \multicolumn{1}{c}{21.2} & 34.4        & \textbf{+13.1} / 62\%\\
CodeLlama-7B& 17.4                     & 27.8        & +10.4 / 60\%\\
CodeLlama-13B& 20.8                     & 29.0        & +8.2 / 39\%\\
CodeLlama-34B& \multicolumn{1}{c}{22.2} & 27.8        & +5.6 / 25\%\\ 
GPT-3.5          & \textbf{31.4}                     & \textbf{39.2}        & +7.8  / 25\%\\
\bottomrule
\end{tabular}
\label{tab:few_vs_ft}
\end{minipage}
\end{table}

\textbf{Fine-tuning significantly increases agent performances}. 
As shown in Table~\ref{tab:few_vs_ft}, fine-tuning consistently and significantly improves the HotpotQA EM from prompting. While weaker LMs benefit more from fine-tuning (e.g., Llama-2-7B increases by 77\%), even strong LMs such as GPT-3.5 could improve performances by 25\%, clearly showing the benefit of learning from more samples.
When compared to strong prompting baselines in Table~\ref{tab:hotpot_baseline}, we find fine-tuned Llama-2-13B could outperform all GPT-3.5 prompting methods (Input-Output prompting, IO; Chain-of-thought, \cott{}; \react{}). It is a promising signal that fine-tuning small open-source LMs could outperform prompting  stronger commercial LMs.
Finally, fine-tuned GPT-3.5, which is the strongest fine-tuned LM, could outperform GPT-4 + IO prompting but still lags behind GPT-4 + \cott{}/\react{} prompting, suggesting room for improvement. More results (e.g., standard error) are in Appendix \ref{app:std}.

\textbf{Fine-tuning is cheaper and faster during agent inference}. 
Since few-shot in-context examples are not needed for fine-tuned LMs, their inference becomes more efficient,  especially  for agentic applications where the context is iteratively accumulated.
For example, the first part of Table~\ref{tab:together} compares costs of fine-tuned vs.\,prompted GPT-3.5 inference, and finds the inference time is reduced by 70\% (9.0s to 2.7s per trial), and the inference cost is reduced even though fine-tuned inference is charged $8\times$ expensive. 
While these costs will vary by conditions (e.g., parallelism implementation), the advantage of having a much smaller context is clear.

\subsection{Robustness and generalization} 

\textbf{Robustness to noisy tools}. The tools or environments that language agents interact with are not always trustworthy, which has led to safety concerns like jailbreaking~\citep{liu2023jailbreaking} or prompt injection~\citep{willison2023injection}. Here we consider a simplified and harmless setup, where the search API has a probability of 0.5 to return 1) ``None'' or 2) a random search response (from all previous experiments and trials), and ask if language agents could still robustly answer questions. As shown in the second part of Table~\ref{tab:together}, the ``None'' setup turns out the more challenging one, which lowered \react{} EM by 33.8\% and \method{} EM only by 14.2\%. Interestingly, random observations hurt \react{} by a similar degree (28.0\% drop), but does not hurt \method{} much (only 5.1\% drop), possibly because the fine-tuning trajectories already contain examples of noisy search queries and how GPT-4 ``reacts'' to such noises successfully. These initial results hint at the importance of a more diverse learning support for robustness. More results on robustness can be found in Appendix~\ref{app:robustness}.

\textbf{Generalization to new tasks}. Table~\ref{tab:together}'s third part shows EM results of fine-tuned and prompted GPT-3.5 on Bamboogle~\citep{press2022measuring}, a test set of 125 multi-hop questions carefully crafted such that searching the questions on Google cannot directly give answers. While HotpotQA fine-tuned or prompted GPT-3.5 both generalize to Bamboogle reasonably, the former (44.0 EM) still beats the latter (40.8 EM), suggesting generalization advantages of fine-tuning. Similarly, combined with the few-shot prompts, fine-tuning on HotpotQA greatly improves the performance on Bamboogle, while slightly improving on MMLU and downgrading on StrategyQA compared to vanilla models (Appendix \ref{app:few_shot_gen}). Since fine-tuning on HotpotQA could hardly generalize to StrategyQA (yes/no questions) or MMLU (multi-choice questions), two other QA datasets with different question styles and answer formats, it motivates our multi-task fine-tuning experiments in Section~\ref{sec:multi-task}.

\begin{table}[t]
\centering
\caption{Comparison of costs, robustness, and generalization for fine-tuned vs. prompted GPT-3.5.}
\begin{tabular}{l|cc|ccc|c}
\toprule
 & \multicolumn{2}{c|}{\textbf{Cost per trial}} & \multicolumn{3}{c|}{\textbf{Obs. Robustness (EM})} & \textbf{Generalization} \\ 
 & Money (\$)             & Time (s)   & Normal     & ``None''     & Random     &  Bamboogle (EM) \\ \hline
\method{}                & $\mathbf{2.2 \times 10^{-3}}$   & \textbf{2.7}        & \textbf{39.2}          & \textbf{33.6}          & \textbf{37.2}          & \textbf{44.0}           \\
\react{}                    & $2.6 \times 10^{-3}$   & 9.0        & 31.4          & 20.8          & 22.6          & 40.8         \\
\bottomrule
\end{tabular}
\label{tab:together}
\end{table}

\begin{figure}[t]
    \centering
    \begin{minipage}{0.3\linewidth}
        \centering
        \includegraphics[width=1\linewidth]{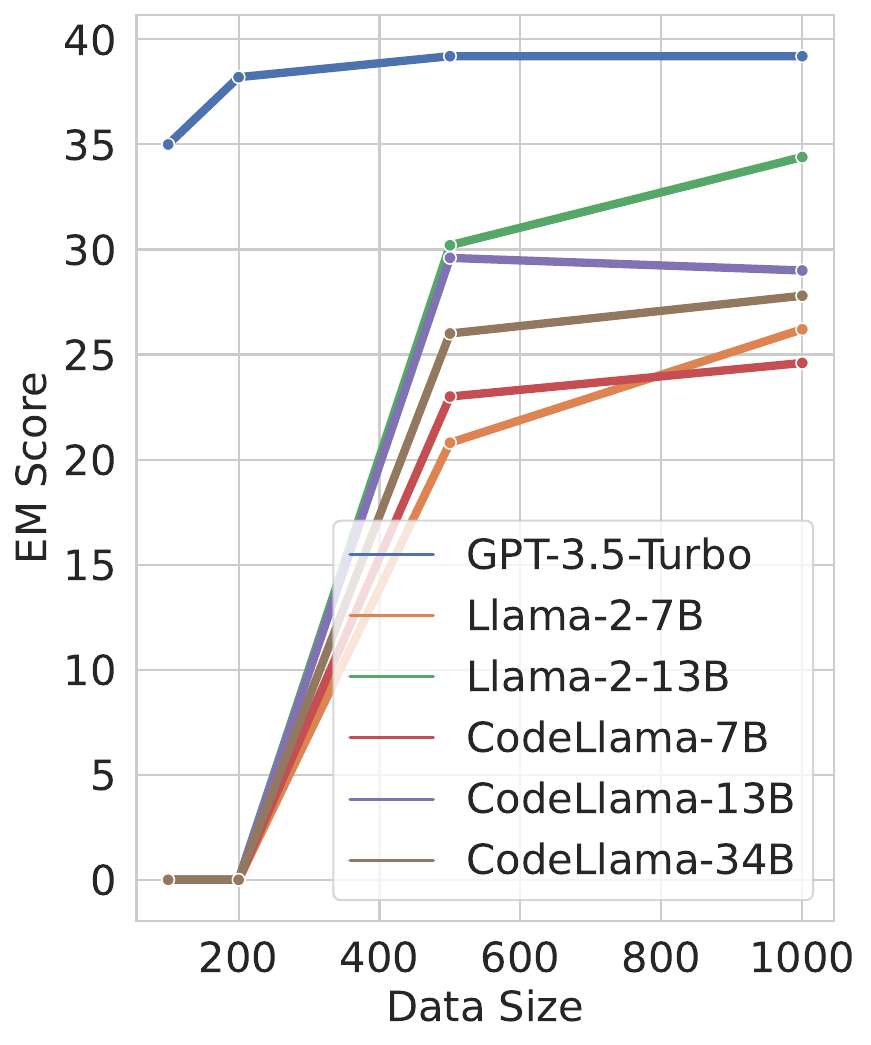}
        \caption{Data scaling.}
        \label{fig:data-scale}
    \end{minipage}%
    \begin{minipage}{0.7\linewidth}
        \centering
        \includegraphics[width=1\linewidth]{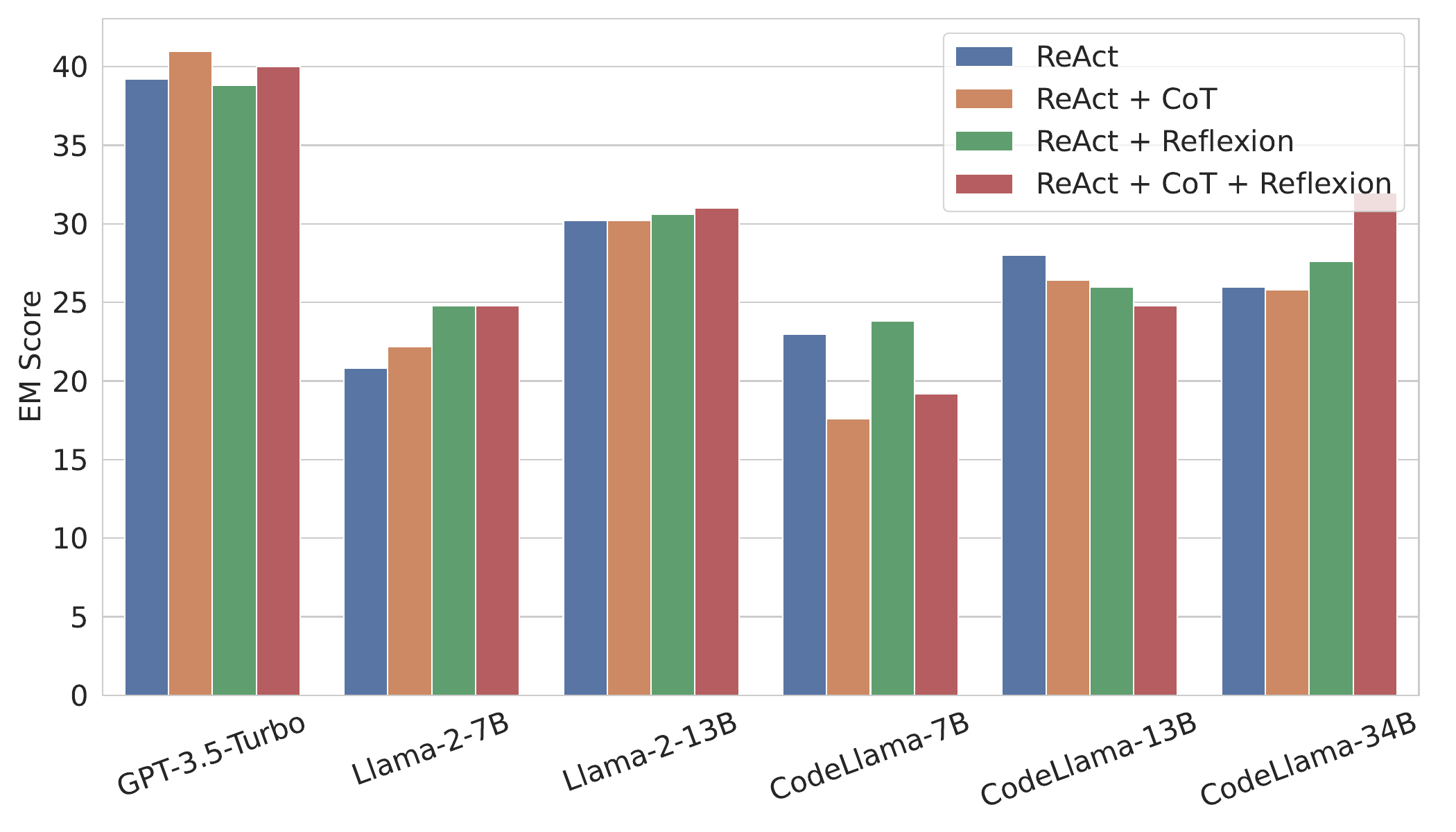}
        \caption{Results across different LMs and data types.}
        \label{fig:data-type}
    \end{minipage}
\end{figure}

\subsection{Analysis of various fine-tuning factors}

\textbf{Effect of fine-tuning method (LoRA vs. Full model)}. For Llama-2-7B, we observe that full-model fine-tuning (30.2 EM) outperforms LoRA fine-tuning (26.2 EM) by 15.3\% (see Appendix~\ref{app:lora_full}). However,  LoRA training is much more affordable, which can train 5.4 examples per second on a single RTX 4090 with 24GB GPU memory, while training 19.7 examples by full fine-tuning requires four A100 GPUs with 80GB GPU memory. Hence, running most experiments with LoRA allows us to explore more training settings with a limited budget and time frame.

\textbf{Effect of fine-tuning data scale}. 
Figure~\ref{fig:data-scale} shows how \method{} performances scale with the number of fine-tuning trajectories ($n \in \{100, 200, 500, 1000\}$). 
GPT-3.5 appears very sample-efficient, requiring only 100 samples to reach an EM around 35, and the gain after 200 samples is marginal. 
On the other hand, Llama models cannot even learn the \react{} format using 100 or 200 samples, but non-trivial scores ``emerge'' with 500 samples, and most models (except CodeLlama-13B) further improve with 1,000 samples. Such a data scaling trend suggests that smaller open-source LMs could potentially catch up with stronger LMs on a particular agentic task given enough fine-tuning data (e.g., Llama-2-13B fine-tuned on 1,000 samples can match  GPT-3.5 fine-tuned on 100 samples).

\textbf{Effect of base LM type}. Table~\ref{tab:few_vs_ft} reveals that GPT-3.5 is superior to all Llama-based models in both prompting and fine-tuning configurations. Additionally, CodeLlama-7B outperforms Llama-2-7B, while CodeLlama-13B does not perform as well as Llama-2-13B, suggesting that coding fine-tuning may not always be beneficial for agentic use cases. CodeLlama performs slightly better when using the default CodeLlama tokenizer instead of the Llama tokenizer (Appendix \ref{app:code_tokenizer}).

\textbf{Effect of base LM scale}. As can be seen in Table~\ref{tab:few_vs_ft} or the blue bars of Figure~\ref{fig:data-type}, (Code)Llama models with 13B parameters always outperform ones with 7B parameters, but CodeLlama-34B seems worse than CodeLlama-13B when fine-tuned purely on \react{} trajectories. But as we will see in Section~\ref{sec:multi-method} (and hinted in the rest of Figure~\ref{fig:data-type}), other factors such as the fine-tuning data type might affect the conclusion and make CodeLlama-34B outperforming CodeLlama-13B. 
In general, multiple components (LM type, LM scale, fine-tuning data and method) might influence fine-tuning results jointly, so different dimensions of scaling trends and  LM/data types should also be considered jointly for agent design. 

\begin{figure}[t]
    \centering
    \includegraphics[width=\linewidth]{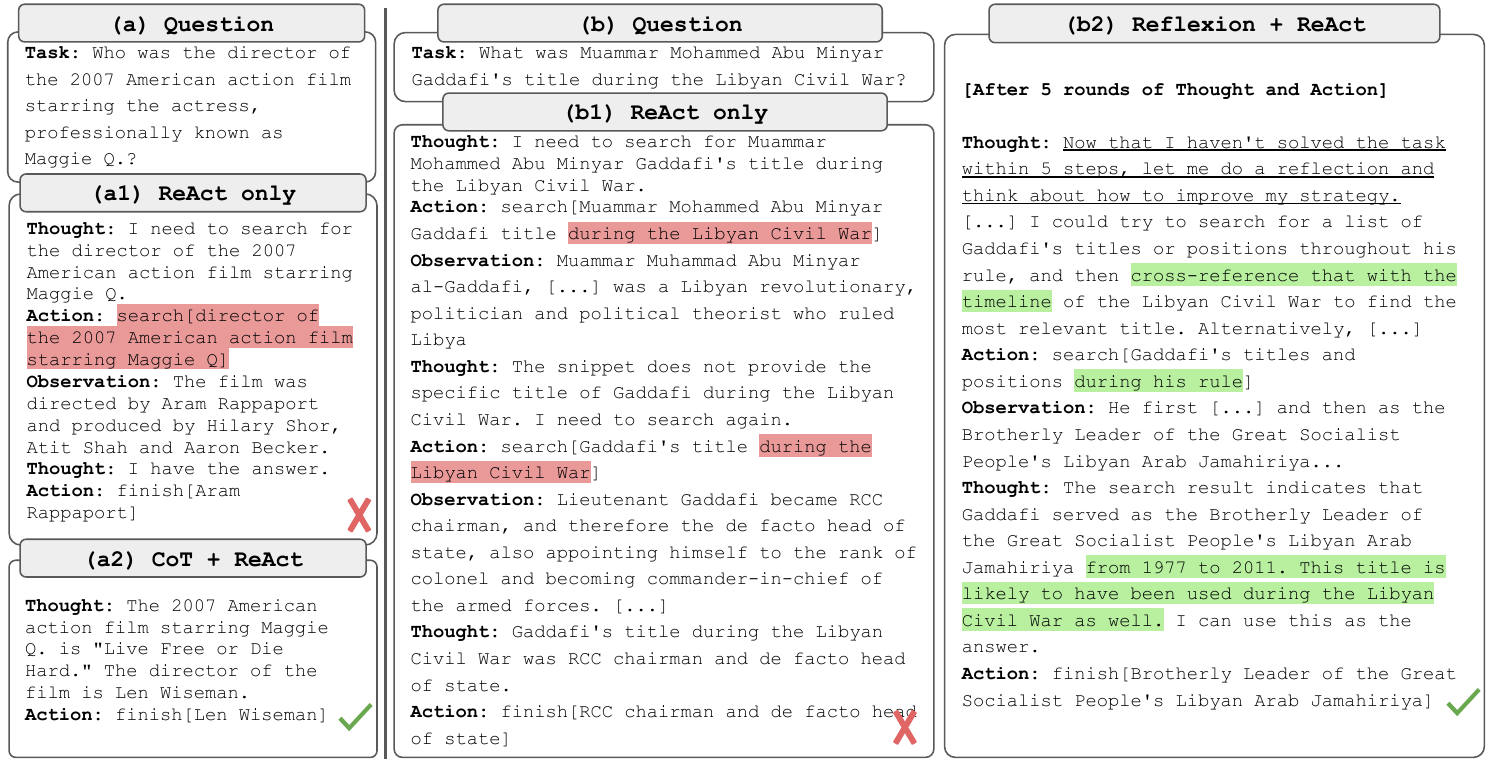}
    \caption{
    \textbf{For question (a)}, (a1) \react{}-only fine-tuning leads to distracting information and wrong answer, while (a2) \cott{} + \react{} fine-tuning leads to a simple \cott{} solution.
    \textbf{For question (b)}, (b1) \react{}-only fine-tuning leads to mutiple failed searches with the same patterns, while (b2) \reflexion{} + \react{} fine-tuning leads to a successful pivot of the search strategy.
    }
    \label{fig:traj}
\end{figure}
\section{Multi-method Fine-tuning}
\label{sec:multi-method}
Next we integrate \cott{}~\citep{wei2022chain} and \reflexion{}~\citep{shinn2023reflexion} with \react{} for multi-method fine-tuning on HotpotQA. For both methods, we generate 500 few-shot prompting trajectories via GPT-4, and use 47 long \reflexion{} trajectories that incorporated self-reflections after 6 or 10 \react{} rounds, and 187 successful \cott{} trajectories reformatted as single-round \react{} trajectories, on top of 500 existing \react{} trajectories. More  details are in Appendix \ref{app:exp_detail}. 

\textbf{Multi-method fine-tuning increases agent flexibility}.
Before quantitative results, we present two example questions in Figure~\ref{fig:traj} and some fine-tuned GPT-3.5 trajectories to illustrate the benefit of multi-method \method{} fine-tuning. The first question (a) is simple, but the \react{}-only fine-tuned agent (a1) searched an over-complicated query that led to the distraction and wrong answer. In contrast, an agent fine-tunend with both CoT and \react{} chose to solve the task within one round relying on confident internal knowledge. The second question (b) is harder, and  the \react{}-only fine-tuned agent (b1) kept searching queries ending in ``during the Libyan Civil War'' without useful information. In contrast, an agent fine-tuned with both Reflexion and \react{} reflected upon this problem, and pivoted the search strategy to change the time constraint to ``during his rule'', which led to the right answer. The flexibility to implicitly choose methods for different problems is another key advantage of fine-tuning over prompting.

\begin{table}[t]

\begin{minipage}{0.49\linewidth}
\centering
\caption{Multi-method results on HotpotQA.}
\resizebox{\columnwidth}{!}{%
\begin{tabular}{lccc} \toprule
\multirow{2}{*}{prompting method}&      & \multicolumn{2}{c}{\#Turns} \\
& EM   & $\mu$           & $\sigma$          \\ \hline
\react{}&      39.4 & 3.2             & 1.4               \\
 \react{} + \cott{}& \textbf{41.0} & 2.7             &1.7               \\
                      \react{} + \reflexion{}& 38.8 & \textbf{3.8}             & 2.8               \\ 
                      \react{} + \cott{} + \reflexion{}& 40.0 & 3.0             & 4.8               \\ \hline
Random method choice & 32.4 & -             & -                 \\ 
Oracle method choice & 52.0 & -             & -               \\ 
                      \bottomrule
\end{tabular}
}
\label{tab:data_mix_study}
\end{minipage}
\begin{minipage}{0.5\linewidth}
\centering
\caption{Multi-task results with GPT-3.5.}
\resizebox{\columnwidth}{!}{%
\begin{tabular}{@{}lcccc@{}}
\toprule
                   & {{HotpotQA}} & {{StrategyQA}}    &  {{Bamboogle}} & {{MMLU}}          \\ \midrule
\textbf{Prompting}  &                              &               &                &\\
IO                 & 22.4                         & 48.0          &  7.2&\textbf{68.6} \\
\cott{}       & 28.0                         & 49.0          &  41.6&50.8          \\
\react{}     & 31.4                         & 61.0          &  40.8&58.6          \\ \midrule
\textbf{Fine-tuning}   &                              &               &                &\\
HotpotQA           & 39.2                         & -             &  44.0&-             \\
Multi-task          & 39.2                         & 55.5          &  43.2&63.2          \\
\hspace{4mm} + \cott{} & \textbf{39.6}                & \textbf{72.9} &  \textbf{50.4}&65.8          \\ \bottomrule
\end{tabular}
}
\label{tab:task_mix_study}
\end{minipage}
\end{table}

\textbf{Multi-method fine-tuning affect different LMs differently}. 
Despite the intuitive benefit, Figure~\ref{fig:data-type} shows mixing more methods does not always improve results, and the optimal mix of methods depends on the base LM. For example, \react{}+\cott{} outperforms \react{} for GPT-3.5 and Llama-2 models, but hurts for CodeLlama models. \react{}+\cott{}+\reflexion{} is the worst for CodeLlama-7/13B, but is the best for CodeLlama-34B. These non-trivial results call for further studies of the interaction of base LMs and fine-tuning data.

\textbf{Can multi-method agents choose suitable methods?}  Table~\ref{tab:data_mix_study} displays HotpotQA test results of various  \method{} agents based on GPT-3.5, as well as the mean ($\mu$) and standard deviation ($\sigma$) of the number of \react{} rounds across their trajectories. Compared to \react{}-only fine-tuning, \react{}+\cott{} improves the EM and reduces the trajectory length, while \react{}+\reflexion{} hurts the EM and increases the trajectory length.  This suggests the two method mixes shift the method selection to two different directions, and \cott{} is perhaps more helpful for HotpotQA questions.
To further understand if multi-method agents could choose the suitable methods, we calculate the result of randomly choosing a method during inference. The result of 32.4 is much lower than all multi-method agents, suggesting the method selection is non-trivial. But applying the best method for each instance leads to an ``oracle'' result of 52.0, suggesting room for improving prompting method selection. Future work could explore more systematic grid search or connections between trajectory statistics and performances to set up better method mix ratios.

\section{Multi-task fine-tuning}
\label{sec:multi-task}

So far fine-tuning has only used HotpotQA data, but empirical studies on LM fine-tuning have shown the benefit of mixing different tasks~\citep{longpre2023flan}. 
Here we fine-tune GPT-3.5 using a mix of training data from three datasets: HotpotQA (500 \react{} samples, 277 \cott{} samples), StrategyQA (388 \react{} samples, 380 \cott{} samples), and MMLU (456 \react{} samples, 469 \cott{} samples). These samples are picked from successful \react{}/\cott{} few-shot prompting trajectories generated via GPT-4.

As shown in Table \ref{tab:task_mix_study}, when StrategyQA/MMLU data is added (``Multi-task''), HotpotQA/Bamboogle performances almost remain the same. 
On one hand, StrategyQA/MMLU trajectories contain very different questions (e.g., MMLU questions are multi-choice) and tool use strategies (e.g., MMLU \react{} trajectories tend to search answer choices), which makes transfer hard.
On the other hand, despite the distribution shift, adding StrategyQA/MMLU does not hurt  HotpotQA/Bamboogle performances, which hints at the promise of fine-tuning one multi-task agent to replace multiple single-task agents, without worrying about negative cross-task influences. 

When we switch from multi-task, single-method fine-tuning to multi-task, multi-method fine-tuning, we find increased performances across all tasks, again reinforcing the value of multi-method agent fine-tuning. Intriguingly, all fine-tuned agents (plus \cott{}/\react{} prompting) underperform naive input-output (IO) prompting on MMLU. One possible explanation is these questions might be too easy to require reasoning and acting, and another explanation could be answer choice memorization. This urges efforts for better prompting methods as well as for better agent datasets.

\section{Discussion}

\textbf{When to fine-tune vs.\,prompt for language agents?} While most existing language agents use prompting, our work calls for a re-thinking of best practices by showing multi-facet benefits of fine-tuning as a result of more diverse learning support. Thus, prompting and fine-tuning seem more suitable for exploration and exploitation usecases respectively. To develop new agents or solve new tasks, prompting off-the-shelf LMs could provide flexibility and convenience. On the other hand, when the downstream task is known (e.g., QA), effective prompting methods for agents have been explored (e.g., ReAct), and enough data can be collected (e.g., via GPT-4), fine-tuning can provide better performances, stronger generalization to new tasks, more robustness to noisy or adversarial environments, as well as cheaper and  more efficient inference.  These features make fine-tuning especially attractive when used for large-scale industrial solutions. 

\textbf{Which LM to fine-tune?} Of all models we considered, GPT-3.5 consistently outperforms other Llama-based LMs in various setups, which is not surprising given its much larger model size and continued training from GPT-3. It also has better sample efficiency and a reasonable cost (around \$10 per fine-tuning experiment in our case). However, we have also shown that open source Llama models could be fine-tuned to catch up with GPT-3.5 performances, given enough fine-tuning data with the right mixing of prompting methods and tasks. Practioners should balance the tradeoff between the convenience and performance of GPT-3.5 versus the controlability and reproducibility of open-source LMs for agent fine-tuning.

\textbf{When to use tools or reflect for language agents?} Prompting-based language agents can only imitate  a small and fixed set of successful task-solving trajectories. This could lead to tool overuse (e.g., search for knowledge already stored in LMs), and inabilities to recover when the trajectory deviates from the ``successful'' patterns (e.g., keep searching similar queries with useless observations). \method{}'s multi-method fine-tuning helps increase a language agent's flexibility and robustness, but the problem of knowing when to get help (tool use) and feedback (reflection) is still far from being solved. Work on calibration~\citep{ren2023robots} and meta-reasoning~\citep{griffiths2019doing} might shed light into better agent designs in this regard.

\textbf{Limitations and future directions}. This work is an initial step toward language agent fine-tuning, and is constrained to a single type of task (QA) and a single tool (Google search). Future work could apply the research questions raised by \method{} to more tasks and grounding setups (e.g., more API tools, the web, the physical world). Also, we focused on three methods (\react{}, \cott{}, \reflexion{}) that maintain a single autoregressive trajectory context, which makes fine-tuning straightforward. It remains underexplored how to fine-tune more advanced agents involving multiple prompts, roles, and contexts~\citep{wang2023voyager, park2023generative, yao2023tree}, or best combine prompting and fine-tuning in a complex agent system. Finally, the multi-task setup in this work is limited to three QA tasks, and the best LM we could fine-tune is GPT-3.5. A large-scale multi-task fine-tuning~\citep{wei2022finetuned} using the state-of-the-art LM backbone will test the limit of language agent fine-tuning, but more suitable and diverse benchmarks to develop and evaluate agents should be explored first.

\section*{Acknowledgements}
We thank Yuqian Sun for figure helps, SerpAPI for funding a part of API calls, and Tianyu Gao, Ofir Press, Noah Shinn, Alex Witegg, Eric Zelikman, and Zexuan Zhong for proofreading and valuable feedback.
SY and KN acknowledge support from an Oracle Collaborative Research award and the National Science Foundation under Grant No. 2239363. SY is also supported by the Harold W. Dodds Fellowship from Princeton. Any opinions, findings, conclusions, or recommendations expressed in this material are those of the author(s) and do not necessarily reflect the views of the National Science Foundation. We are also grateful to acknowledge that the work of the joint first author, CS, has been jointly supported by a donation from Toshiba Europe and the Engineering and Physical Sciences Research Council of UKRI (grant number 2752931).

\bibliography{main}

\begin{thebibliography}{81}
\providecommand{\natexlab}[1]{#1}
\providecommand{\url}[1]{\texttt{#1}}
\expandafter\ifx\csname urlstyle\endcsname\relax
  \providecommand{\doi}[1]{doi: #1}\else
  \providecommand{\doi}{doi: \begingroup \urlstyle{rm}\Url}\fi

\bibitem[Ahn et~al.(2022)Ahn, Brohan, Brown, Chebotar, Cortes, David, Finn, Fu, Gopalakrishnan, Hausman, et~al.]{ahn2022can}
Michael Ahn, Anthony Brohan, Noah Brown, Yevgen Chebotar, Omar Cortes, Byron David, Chelsea Finn, Chuyuan Fu, Keerthana Gopalakrishnan, Karol Hausman, et~al.
\newblock Do as {I} can, not as {I} say: Grounding language in robotic affordances.
\newblock \emph{arXiv preprint arXiv:2204.01691}, 2022.

\bibitem[Bharadhwaj et~al.(2023)Bharadhwaj, Vakil, Sharma, Gupta, Tulsiani, and Kumar]{DBLP:journals/corr/abs-2309-01918}
Homanga Bharadhwaj, Jay Vakil, Mohit Sharma, Abhinav Gupta, Shubham Tulsiani, and Vikash Kumar.
\newblock Roboagent: Generalization and efficiency in robot manipulation via semantic augmentations and action chunking.
\newblock \emph{CoRR}, abs/2309.01918, 2023.
\newblock \doi{10.48550/arXiv.2309.01918}.
\newblock URL \url{https://doi.org/10.48550/arXiv.2309.01918}.

\bibitem[Brooks et~al.(2023)Brooks, Holynski, and Efros]{DBLP:conf/cvpr/BrooksHE23}
Tim Brooks, Aleksander Holynski, and Alexei~A. Efros.
\newblock Instructpix2pix: Learning to follow image editing instructions.
\newblock In \emph{{IEEE/CVF} Conference on Computer Vision and Pattern Recognition, {CVPR} 2023, Vancouver, BC, Canada, June 17-24, 2023}, pp.\  18392--18402. {IEEE}, 2023.
\newblock \doi{10.1109/CVPR52729.2023.01764}.
\newblock URL \url{https://doi.org/10.1109/CVPR52729.2023.01764}.

\bibitem[Brown et~al.(2020)Brown, Mann, Ryder, Subbiah, Kaplan, Dhariwal, Neelakantan, Shyam, Sastry, Askell, et~al.]{brown2020language}
Tom Brown, Benjamin Mann, Nick Ryder, Melanie Subbiah, Jared~D Kaplan, Prafulla Dhariwal, Arvind Neelakantan, Pranav Shyam, Girish Sastry, Amanda Askell, et~al.
\newblock Language models are few-shot learners.
\newblock \emph{Advances in Neural Information Processing Systems}, 33:\penalty0 1877--1901, 2020.

\bibitem[Chiang et~al.(2023)Chiang, Li, Lin, Sheng, Wu, Zhang, Zheng, Zhuang, Zhuang, Gonzalez, Stoica, and Xing]{vicuna2023}
Wei-Lin Chiang, Zhuohan Li, Zi~Lin, Ying Sheng, Zhanghao Wu, Hao Zhang, Lianmin Zheng, Siyuan Zhuang, Yonghao Zhuang, Joseph~E. Gonzalez, Ion Stoica, and Eric~P. Xing.
\newblock Vicuna: An open-source chatbot impressing gpt-4 with 90\%* chatgpt quality, March 2023.
\newblock URL \url{https://lmsys.org/blog/2023-03-30-vicuna/}.

\bibitem[Chowdhery et~al.(2022)Chowdhery, Narang, Devlin, Bosma, Mishra, Roberts, Barham, Chung, Sutton, Gehrmann, et~al.]{chowdhery2022palm}
Aakanksha Chowdhery, Sharan Narang, Jacob Devlin, Maarten Bosma, Gaurav Mishra, Adam Roberts, Paul Barham, Hyung~Won Chung, Charles Sutton, Sebastian Gehrmann, et~al.
\newblock Palm: Scaling language modeling with pathways.
\newblock \emph{arXiv preprint arXiv:2204.02311}, 2022.

\bibitem[Dagan et~al.(2023)Dagan, Keller, and Lascarides]{dagan2023dynamic}
Gautier Dagan, Frank Keller, and Alex Lascarides.
\newblock Dynamic planning with a llm.
\newblock \emph{arXiv preprint arXiv:2308.06391}, 2023.

\bibitem[Dai et~al.(2023)Dai, Li, Li, Tiong, Zhao, Wang, Li, Fung, and Hoi]{DBLP:journals/corr/abs-2305-06500}
Wenliang Dai, Junnan Li, Dongxu Li, Anthony Meng~Huat Tiong, Junqi Zhao, Weisheng Wang, Boyang Li, Pascale Fung, and Steven C.~H. Hoi.
\newblock Instructblip: Towards general-purpose vision-language models with instruction tuning.
\newblock \emph{CoRR}, abs/2305.06500, 2023.
\newblock \doi{10.48550/arXiv.2305.06500}.
\newblock URL \url{https://doi.org/10.48550/arXiv.2305.06500}.

\bibitem[Deng et~al.(2023)Deng, Gu, Zheng, Chen, Stevens, Wang, Sun, and Su]{deng2023mind2web}
Xiang Deng, Yu~Gu, Boyuan Zheng, Shijie Chen, Samuel Stevens, Boshi Wang, Huan Sun, and Yu~Su.
\newblock Mind2{W}eb: Towards a generalist agent for the web.
\newblock \emph{arXiv preprint arXiv:2306.06070}, 2023.

\bibitem[Dettmers et~al.(2023)Dettmers, Pagnoni, Holtzman, and Zettlemoyer]{DBLP:journals/corr/abs-2305-14314}
Tim Dettmers, Artidoro Pagnoni, Ari Holtzman, and Luke Zettlemoyer.
\newblock Qlora: Efficient finetuning of quantized llms.
\newblock \emph{CoRR}, abs/2305.14314, 2023.
\newblock \doi{10.48550/arXiv.2305.14314}.
\newblock URL \url{https://doi.org/10.48550/arXiv.2305.14314}.

\bibitem[Ding et~al.(2023)Ding, Qin, Yang, Wei, Yang, Su, Hu, Chen, Chan, Chen, Yi, Zhao, Wang, Liu, Zheng, Chen, Liu, Tang, Li, and Sun]{DBLP:journals/natmi/DingQYWYSHCCCYZWLZCLTLS23}
Ning Ding, Yujia Qin, Guang Yang, Fuchao Wei, Zonghan Yang, Yusheng Su, Shengding Hu, Yulin Chen, Chi{-}Min Chan, Weize Chen, Jing Yi, Weilin Zhao, Xiaozhi Wang, Zhiyuan Liu, Hai{-}Tao Zheng, Jianfei Chen, Yang Liu, Jie Tang, Juanzi Li, and Maosong Sun.
\newblock Parameter-efficient fine-tuning of large-scale pre-trained language models.
\newblock \emph{Nat. Mac. Intell.}, 5\penalty0 (3):\penalty0 220--235, 2023.
\newblock \doi{10.1038/s42256-023-00626-4}.
\newblock URL \url{https://doi.org/10.1038/s42256-023-00626-4}.

\bibitem[Driess et~al.(2023)Driess, Xia, Sajjadi, Lynch, Chowdhery, Ichter, Wahid, Tompson, Vuong, Yu, Huang, Chebotar, Sermanet, Duckworth, Levine, Vanhoucke, Hausman, Toussaint, Greff, Zeng, Mordatch, and Florence]{DBLP:conf/icml/DriessXSLCIWTVY23}
Danny Driess, Fei Xia, Mehdi S.~M. Sajjadi, Corey Lynch, Aakanksha Chowdhery, Brian Ichter, Ayzaan Wahid, Jonathan Tompson, Quan Vuong, Tianhe Yu, Wenlong Huang, Yevgen Chebotar, Pierre Sermanet, Daniel Duckworth, Sergey Levine, Vincent Vanhoucke, Karol Hausman, Marc Toussaint, Klaus Greff, Andy Zeng, Igor Mordatch, and Pete Florence.
\newblock Palm-e: An embodied multimodal language model.
\newblock In Andreas Krause, Emma Brunskill, Kyunghyun Cho, Barbara Engelhardt, Sivan Sabato, and Jonathan Scarlett (eds.), \emph{International Conference on Machine Learning, {ICML} 2023, 23-29 July 2023, Honolulu, Hawaii, {USA}}, volume 202 of \emph{Proceedings of Machine Learning Research}, pp.\  8469--8488. {PMLR}, 2023.
\newblock URL \url{https://proceedings.mlr.press/v202/driess23a.html}.

\bibitem[Geva et~al.(2021)Geva, Khashabi, Segal, Khot, Roth, and Berant]{DBLP:journals/tacl/GevaKSKRB21}
Mor Geva, Daniel Khashabi, Elad Segal, Tushar Khot, Dan Roth, and Jonathan Berant.
\newblock Did aristotle use a laptop? {A} question answering benchmark with implicit reasoning strategies.
\newblock \emph{Trans. Assoc. Comput. Linguistics}, 9:\penalty0 346--361, 2021.
\newblock \doi{10.1162/tacl\_a\_00370}.
\newblock URL \url{https://doi.org/10.1162/tacl\_a\_00370}.

\bibitem[Gong et~al.(2023)Gong, Lyu, Zhang, Wang, Zheng, Zhao, Liu, Zhang, Luo, and Chen]{DBLP:journals/corr/abs-2305-04790}
Tao Gong, Chengqi Lyu, Shilong Zhang, Yudong Wang, Miao Zheng, Qian Zhao, Kuikun Liu, Wenwei Zhang, Ping Luo, and Kai Chen.
\newblock Multimodal-gpt: {A} vision and language model for dialogue with humans.
\newblock \emph{CoRR}, abs/2305.04790, 2023.
\newblock \doi{10.48550/arXiv.2305.04790}.
\newblock URL \url{https://doi.org/10.48550/arXiv.2305.04790}.

\bibitem[Griffiths et~al.(2019)Griffiths, Callaway, Chang, Grant, Krueger, and Lieder]{griffiths2019doing}
Thomas~L Griffiths, Frederick Callaway, Michael~B Chang, Erin Grant, Paul~M Krueger, and Falk Lieder.
\newblock Doing more with less: meta-reasoning and meta-learning in humans and machines.
\newblock \emph{Current Opinion in Behavioral Sciences}, 29:\penalty0 24--30, 2019.

\bibitem[Gunasekar et~al.(2023)Gunasekar, Zhang, Aneja, Mendes, Giorno, Gopi, Javaheripi, Kauffmann, de~Rosa, Saarikivi, Salim, Shah, Behl, Wang, Bubeck, Eldan, Kalai, Lee, and Li]{gunasekar2023textbooks}
Suriya Gunasekar, Yi~Zhang, Jyoti Aneja, Caio César~Teodoro Mendes, Allie~Del Giorno, Sivakanth Gopi, Mojan Javaheripi, Piero Kauffmann, Gustavo de~Rosa, Olli Saarikivi, Adil Salim, Shital Shah, Harkirat~Singh Behl, Xin Wang, Sébastien Bubeck, Ronen Eldan, Adam~Tauman Kalai, Yin~Tat Lee, and Yuanzhi Li.
\newblock Textbooks are all you need, 2023.

\bibitem[Guu et~al.(2020)Guu, Lee, Tung, Pasupat, and Chang]{DBLP:conf/icml/GuuLTPC20}
Kelvin Guu, Kenton Lee, Zora Tung, Panupong Pasupat, and Ming{-}Wei Chang.
\newblock Retrieval augmented language model pre-training.
\newblock In \emph{Proceedings of the 37th International Conference on Machine Learning, {ICML} 2020, 13-18 July 2020, Virtual Event}, volume 119 of \emph{Proceedings of Machine Learning Research}, pp.\  3929--3938. {PMLR}, 2020.
\newblock URL \url{http://proceedings.mlr.press/v119/guu20a.html}.

\bibitem[Hendrycks et~al.(2021)Hendrycks, Burns, Basart, Zou, Mazeika, Song, and Steinhardt]{DBLP:conf/iclr/HendrycksBBZMSS21}
Dan Hendrycks, Collin Burns, Steven Basart, Andy Zou, Mantas Mazeika, Dawn Song, and Jacob Steinhardt.
\newblock Measuring massive multitask language understanding.
\newblock In \emph{9th International Conference on Learning Representations, {ICLR} 2021, Virtual Event, Austria, May 3-7, 2021}. OpenReview.net, 2021.
\newblock URL \url{https://openreview.net/forum?id=d7KBjmI3GmQ}.

\bibitem[Hinton et~al.(2015)Hinton, Vinyals, and Dean]{hinton2015distilling}
Geoffrey Hinton, Oriol Vinyals, and Jeff Dean.
\newblock Distilling the knowledge in a neural network.
\newblock \emph{arXiv preprint arXiv:1503.02531}, 2015.

\bibitem[Honovich et~al.(2023)Honovich, Scialom, Levy, and Schick]{DBLP:conf/acl/HonovichSLS23}
Or~Honovich, Thomas Scialom, Omer Levy, and Timo Schick.
\newblock Unnatural instructions: Tuning language models with (almost) no human labor.
\newblock In Anna Rogers, Jordan~L. Boyd{-}Graber, and Naoaki Okazaki (eds.), \emph{Proceedings of the 61st Annual Meeting of the Association for Computational Linguistics (Volume 1: Long Papers), {ACL} 2023, Toronto, Canada, July 9-14, 2023}, pp.\  14409--14428. Association for Computational Linguistics, 2023.
\newblock \doi{10.18653/v1/2023.acl-long.806}.
\newblock URL \url{https://doi.org/10.18653/v1/2023.acl-long.806}.

\bibitem[Hu et~al.(2022)Hu, Shen, Wallis, Allen{-}Zhu, Li, Wang, Wang, and Chen]{DBLP:conf/iclr/HuSWALWWC22}
Edward~J. Hu, Yelong Shen, Phillip Wallis, Zeyuan Allen{-}Zhu, Yuanzhi Li, Shean Wang, Lu~Wang, and Weizhu Chen.
\newblock Lora: Low-rank adaptation of large language models.
\newblock In \emph{The Tenth International Conference on Learning Representations, {ICLR} 2022, Virtual Event, April 25-29, 2022}. OpenReview.net, 2022.
\newblock URL \url{https://openreview.net/forum?id=nZeVKeeFYf9}.

\bibitem[Huang et~al.(2022{\natexlab{a}})Huang, Gu, Hou, Wu, Wang, Yu, and Han]{huang2022large}
Jiaxin Huang, Shixiang~Shane Gu, Le~Hou, Yuexin Wu, Xuezhi Wang, Hongkun Yu, and Jiawei Han.
\newblock Large language models can self-improve.
\newblock \emph{arXiv preprint arXiv:2210.11610}, 2022{\natexlab{a}}.

\bibitem[Huang et~al.(2022{\natexlab{b}})Huang, Abbeel, Pathak, and Mordatch]{huang2022language}
Wenlong Huang, Pieter Abbeel, Deepak Pathak, and Igor Mordatch.
\newblock Language models as zero-shot planners: Extracting actionable knowledge for embodied agents.
\newblock In \emph{International Conference on Machine Learning}, pp.\  9118--9147, 2022{\natexlab{b}}.

\bibitem[Ivison et~al.(2023)Ivison, Bhagia, Wang, Hajishirzi, and Peters]{DBLP:conf/acl/IvisonBWHP23}
Hamish Ivison, Akshita Bhagia, Yizhong Wang, Hannaneh Hajishirzi, and Matthew~E. Peters.
\newblock {HINT:} hypernetwork instruction tuning for efficient zero- and few-shot generalisation.
\newblock In Anna Rogers, Jordan~L. Boyd{-}Graber, and Naoaki Okazaki (eds.), \emph{Proceedings of the 61st Annual Meeting of the Association for Computational Linguistics (Volume 1: Long Papers), {ACL} 2023, Toronto, Canada, July 9-14, 2023}, pp.\  11272--11288. Association for Computational Linguistics, 2023.
\newblock \doi{10.18653/v1/2023.acl-long.631}.
\newblock URL \url{https://doi.org/10.18653/v1/2023.acl-long.631}.

\bibitem[Kojima et~al.(2022)Kojima, Gu, Reid, Matsuo, and Iwasawa]{DBLP:conf/nips/KojimaGRMI22}
Takeshi Kojima, Shixiang~Shane Gu, Machel Reid, Yutaka Matsuo, and Yusuke Iwasawa.
\newblock Large language models are zero-shot reasoners.
\newblock In \emph{NeurIPS}, 2022.
\newblock URL \url{http://papers.nips.cc/paper\_files/paper/2022/hash/8bb0d291acd4acf06ef112099c16f326-Abstract-Conference.html}.

\bibitem[K{\"{o}}pf et~al.(2023)K{\"{o}}pf, Kilcher, von R{\"{u}}tte, Anagnostidis, Tam, Stevens, Barhoum, Duc, Stanley, Nagyfi, ES, Suri, Glushkov, Dantuluri, Maguire, Schuhmann, Nguyen, and Mattick]{DBLP:journals/corr/abs-2304-07327}
Andreas K{\"{o}}pf, Yannic Kilcher, Dimitri von R{\"{u}}tte, Sotiris Anagnostidis, Zhi{-}Rui Tam, Keith Stevens, Abdullah Barhoum, Nguyen~Minh Duc, Oliver Stanley, Rich{\'{a}}rd Nagyfi, Shahul ES, Sameer Suri, David Glushkov, Arnav Dantuluri, Andrew Maguire, Christoph Schuhmann, Huu Nguyen, and Alexander Mattick.
\newblock Openassistant conversations - democratizing large language model alignment.
\newblock \emph{CoRR}, abs/2304.07327, 2023.
\newblock \doi{10.48550/arXiv.2304.07327}.
\newblock URL \url{https://doi.org/10.48550/arXiv.2304.07327}.

\bibitem[Le et~al.(2022)Le, Wang, Gotmare, Savarese, and Hoi]{le2022coderl}
Hung Le, Yue Wang, Akhilesh~Deepak Gotmare, Silvio Savarese, and Steven Chu~Hong Hoi.
\newblock Coderl: Mastering code generation through pretrained models and deep reinforcement learning.
\newblock \emph{Advances in Neural Information Processing Systems}, 35:\penalty0 21314--21328, 2022.

\bibitem[Li et~al.(2023)Li, Allal, Zi, Muennighoff, Kocetkov, Mou, Marone, Akiki, Li, Chim, Liu, Zheltonozhskii, Zhuo, Wang, Dehaene, Davaadorj, Lamy-Poirier, Monteiro, Shliazhko, Gontier, Meade, Zebaze, Yee, Umapathi, Zhu, Lipkin, Oblokulov, Wang, Murthy, Stillerman, Patel, Abulkhanov, Zocca, Dey, Zhang, Fahmy, Bhattacharyya, Yu, Singh, Luccioni, Villegas, Kunakov, Zhdanov, Romero, Lee, Timor, Ding, Schlesinger, Schoelkopf, Ebert, Dao, Mishra, Gu, Robinson, Anderson, Dolan-Gavitt, Contractor, Reddy, Fried, Bahdanau, Jernite, Ferrandis, Hughes, Wolf, Guha, von Werra, and de~Vries]{Li2023StarCoderMT}
Raymond Li, Loubna~Ben Allal, Yangtian Zi, Niklas Muennighoff, Denis Kocetkov, Chenghao Mou, Marc Marone, Christopher Akiki, Jia Li, Jenny Chim, Qian Liu, Evgenii Zheltonozhskii, Terry~Yue Zhuo, Thomas Wang, Olivier Dehaene, Mishig Davaadorj, Joel Lamy-Poirier, Jo{\~a}o Monteiro, Oleh Shliazhko, Nicolas Gontier, Nicholas Meade, Armel Zebaze, Ming-Ho Yee, Logesh~Kumar Umapathi, Jian Zhu, Benjamin Lipkin, Muhtasham Oblokulov, Zhiruo Wang, Rudra Murthy, Jason Stillerman, Siva~Sankalp Patel, Dmitry Abulkhanov, Marco Zocca, Manan Dey, Zhihan Zhang, Nourhan Fahmy, Urvashi Bhattacharyya, W.~Yu, Swayam Singh, Sasha Luccioni, Paulo Villegas, Maxim Kunakov, Fedor Zhdanov, Manuel Romero, Tony Lee, Nadav Timor, Jennifer Ding, Claire Schlesinger, Hailey Schoelkopf, Jana Ebert, Tri Dao, Mayank Mishra, Alexander Gu, Jennifer Robinson, Carolyn~Jane Anderson, Brendan Dolan-Gavitt, Danish Contractor, Siva Reddy, Daniel Fried, Dzmitry Bahdanau, Yacine Jernite, Carlos~Mu{\~n}oz Ferrandis, Sean~M. Hughes, Thomas Wolf, Arjun Guha,
  Leandro von Werra, and Harm de~Vries.
\newblock Starcoder: may the source be with you!
\newblock \emph{ArXiv}, abs/2305.06161, 2023.

\bibitem[Liang et~al.(2023)Liang, Huang, Xia, Xu, Hausman, Ichter, Florence, and Zeng]{liang2023code}
Jacky Liang, Wenlong Huang, Fei Xia, Peng Xu, Karol Hausman, Brian Ichter, Pete Florence, and Andy Zeng.
\newblock Code as policies: Language model programs for embodied control.
\newblock In \emph{2023 IEEE International Conference on Robotics and Automation (ICRA)}, pp.\  9493--9500, 2023.

\bibitem[Lin et~al.(2023)Lin, Fu, Yang, Ammanabrolu, Brahman, Huang, Bhagavatula, Choi, and Ren]{lin2023swiftsage}
Bill~Yuchen Lin, Yicheng Fu, Karina Yang, Prithviraj Ammanabrolu, Faeze Brahman, Shiyu Huang, Chandra Bhagavatula, Yejin Choi, and Xiang Ren.
\newblock Swiftsage: A generative agent with fast and slow thinking for complex interactive tasks.
\newblock \emph{arXiv preprint arXiv:2305.17390}, 2023.

\bibitem[Liu et~al.(2023{\natexlab{a}})Liu, Jiang, Zhang, Liu, Zhang, Biswas, and Stone]{liu2023llm+}
Bo~Liu, Yuqian Jiang, Xiaohan Zhang, Qiang Liu, Shiqi Zhang, Joydeep Biswas, and Peter Stone.
\newblock Llm+ p: Empowering large language models with optimal planning proficiency.
\newblock \emph{arXiv preprint arXiv:2304.11477}, 2023{\natexlab{a}}.

\bibitem[Liu et~al.(2023{\natexlab{b}})Liu, Deng, Xu, Li, Zheng, Zhang, Zhao, Zhang, and Liu]{liu2023jailbreaking}
Yi~Liu, Gelei Deng, Zhengzi Xu, Yuekang Li, Yaowen Zheng, Ying Zhang, Lida Zhao, Tianwei Zhang, and Yang Liu.
\newblock Jailbreaking chatgpt via prompt engineering: An empirical study, 2023{\natexlab{b}}.

\bibitem[Longpre et~al.(2023)Longpre, Hou, Vu, Webson, Chung, Tay, Zhou, Le, Zoph, Wei, et~al.]{longpre2023flan}
Shayne Longpre, Le~Hou, Tu~Vu, Albert Webson, Hyung~Won Chung, Yi~Tay, Denny Zhou, Quoc~V Le, Barret Zoph, Jason Wei, et~al.
\newblock The flan collection: Designing data and methods for effective instruction tuning.
\newblock \emph{arXiv preprint arXiv:2301.13688}, 2023.

\bibitem[Luo et~al.(2023)Luo, Xu, Zhao, Sun, Geng, Hu, Tao, Ma, Lin, and Jiang]{DBLP:journals/corr/abs-2306-08568}
Ziyang Luo, Can Xu, Pu~Zhao, Qingfeng Sun, Xiubo Geng, Wenxiang Hu, Chongyang Tao, Jing Ma, Qingwei Lin, and Daxin Jiang.
\newblock Wizardcoder: Empowering code large language models with evol-instruct.
\newblock \emph{CoRR}, abs/2306.08568, 2023.
\newblock \doi{10.48550/arXiv.2306.08568}.
\newblock URL \url{https://doi.org/10.48550/arXiv.2306.08568}.

\bibitem[Lv et~al.(2023)Lv, Yang, Liu, Gao, Guo, and Qiu]{DBLP:journals/corr/abs-2306-09782}
Kai Lv, Yuqing Yang, Tengxiao Liu, Qinghui Gao, Qipeng Guo, and Xipeng Qiu.
\newblock Full parameter fine-tuning for large language models with limited resources.
\newblock \emph{CoRR}, abs/2306.09782, 2023.
\newblock \doi{10.48550/arXiv.2306.09782}.
\newblock URL \url{https://doi.org/10.48550/arXiv.2306.09782}.

\bibitem[Mishra et~al.(2022)Mishra, Khashabi, Baral, and Hajishirzi]{DBLP:conf/acl/MishraKBH22}
Swaroop Mishra, Daniel Khashabi, Chitta Baral, and Hannaneh Hajishirzi.
\newblock Cross-task generalization via natural language crowdsourcing instructions.
\newblock In Smaranda Muresan, Preslav Nakov, and Aline Villavicencio (eds.), \emph{Proceedings of the 60th Annual Meeting of the Association for Computational Linguistics (Volume 1: Long Papers), {ACL} 2022, Dublin, Ireland, May 22-27, 2022}, pp.\  3470--3487. Association for Computational Linguistics, 2022.
\newblock \doi{10.18653/v1/2022.acl-long.244}.
\newblock URL \url{https://doi.org/10.18653/v1/2022.acl-long.244}.

\bibitem[Muennighoff et~al.(2023)Muennighoff, Wang, Sutawika, Roberts, Biderman, Scao, Bari, Shen, Yong, Schoelkopf, Tang, Radev, Aji, Almubarak, Albanie, Alyafeai, Webson, Raff, and Raffel]{DBLP:conf/acl/MuennighoffWSRB23}
Niklas Muennighoff, Thomas Wang, Lintang Sutawika, Adam Roberts, Stella Biderman, Teven~Le Scao, M.~Saiful Bari, Sheng Shen, Zheng~Xin Yong, Hailey Schoelkopf, Xiangru Tang, Dragomir Radev, Alham~Fikri Aji, Khalid Almubarak, Samuel Albanie, Zaid Alyafeai, Albert Webson, Edward Raff, and Colin Raffel.
\newblock Crosslingual generalization through multitask finetuning.
\newblock In Anna Rogers, Jordan~L. Boyd{-}Graber, and Naoaki Okazaki (eds.), \emph{Proceedings of the 61st Annual Meeting of the Association for Computational Linguistics (Volume 1: Long Papers), {ACL} 2023, Toronto, Canada, July 9-14, 2023}, pp.\  15991--16111. Association for Computational Linguistics, 2023.
\newblock \doi{10.18653/v1/2023.acl-long.891}.
\newblock URL \url{https://doi.org/10.18653/v1/2023.acl-long.891}.

\bibitem[Nakano et~al.(2021)Nakano, Hilton, Balaji, Wu, Ouyang, Kim, Hesse, Jain, Kosaraju, Saunders, et~al.]{nakano2021webgpt}
Reiichiro Nakano, Jacob Hilton, Suchir Balaji, Jeff Wu, Long Ouyang, Christina Kim, Christopher Hesse, Shantanu Jain, Vineet Kosaraju, William Saunders, et~al.
\newblock {WebGPT: Browser-Assisted Question-Answering with Human Feedback}.
\newblock \emph{arXiv preprint arXiv:2112.09332}, 2021.

\bibitem[Nye et~al.(2021)Nye, Andreassen, Gur-Ari, Michalewski, Austin, Bieber, Dohan, Lewkowycz, Bosma, Luan, Sutton, and Odena]{nye2021work}
Maxwell Nye, Anders~Johan Andreassen, Guy Gur-Ari, Henryk Michalewski, Jacob Austin, David Bieber, David Dohan, Aitor Lewkowycz, Maarten Bosma, David Luan, Charles Sutton, and Augustus Odena.
\newblock Show your work: Scratchpads for intermediate computation with language models, 2021.

\bibitem[OpenAI(2023{\natexlab{a}})]{openai2023finetune}
OpenAI.
\newblock Gpt-3.5 turbo fine-tuning and api updates, 2023{\natexlab{a}}.
\newblock URL \url{https://openai.com/blog/gpt-3-5-turbo-fine-tuning-and-api-updates}.

\bibitem[OpenAI(2023{\natexlab{b}})]{openai2023gpt4}
OpenAI.
\newblock Gpt-4 technical report, 2023{\natexlab{b}}.

\bibitem[OpenAI(2023{\natexlab{c}})]{plugins}
OpenAI.
\newblock Chatgpt plugins, 2023{\natexlab{c}}.
\newblock URL \url{https://openai.com/blog/chatgpt-plugins}.

\bibitem[Ouyang et~al.(2022)Ouyang, Wu, Jiang, Almeida, Wainwright, Mishkin, Zhang, Agarwal, Slama, Ray, Schulman, Hilton, Kelton, Miller, Simens, Askell, Welinder, Christiano, Leike, and Lowe]{DBLP:conf/nips/Ouyang0JAWMZASR22}
Long Ouyang, Jeffrey Wu, Xu~Jiang, Diogo Almeida, Carroll~L. Wainwright, Pamela Mishkin, Chong Zhang, Sandhini Agarwal, Katarina Slama, Alex Ray, John Schulman, Jacob Hilton, Fraser Kelton, Luke Miller, Maddie Simens, Amanda Askell, Peter Welinder, Paul~F. Christiano, Jan Leike, and Ryan Lowe.
\newblock Training language models to follow instructions with human feedback.
\newblock In \emph{NeurIPS}, 2022.
\newblock URL \url{http://papers.nips.cc/paper\_files/paper/2022/hash/b1efde53be364a73914f58805a001731-Abstract-Conference.html}.

\bibitem[Park et~al.(2023)Park, O'Brien, Cai, Morris, Liang, and Bernstein]{park2023generative}
Joon~Sung Park, Joseph~C O'Brien, Carrie~J Cai, Meredith~Ringel Morris, Percy Liang, and Michael~S Bernstein.
\newblock Generative agents: Interactive simulacra of human behavior.
\newblock \emph{arXiv preprint arXiv:2304.03442}, 2023.

\bibitem[Patil et~al.(2023)Patil, Zhang, Wang, and Gonzalez]{patil2023gorilla}
Shishir~G. Patil, Tianjun Zhang, Xin Wang, and Joseph~E. Gonzalez.
\newblock Gorilla: Large language model connected with massive apis.
\newblock \emph{arXiv preprint arXiv:2305.15334}, 2023.

\bibitem[Press et~al.(2022)Press, Zhang, Min, Schmidt, Smith, and Lewis]{press2022measuring}
Ofir Press, Muru Zhang, Sewon Min, Ludwig Schmidt, Noah~A Smith, and Mike Lewis.
\newblock Measuring and narrowing the compositionality gap in language models.
\newblock \emph{arXiv preprint arXiv:2210.03350}, 2022.

\bibitem[Qin et~al.(2023)Qin, Liang, Ye, Zhu, Yan, Lu, Lin, Cong, Tang, Qian, et~al.]{qin2023toolllm}
Yujia Qin, Shihao Liang, Yining Ye, Kunlun Zhu, Lan Yan, Yaxi Lu, Yankai Lin, Xin Cong, Xiangru Tang, Bill Qian, et~al.
\newblock Toolllm: Facilitating large language models to master 16000+ real-world apis.
\newblock \emph{arXiv preprint arXiv:2307.16789}, 2023.

\bibitem[Ren et~al.(2023)Ren, Dixit, Bodrova, Singh, Tu, Brown, Xu, Takayama, Xia, Varley, et~al.]{ren2023robots}
Allen~Z Ren, Anushri Dixit, Alexandra Bodrova, Sumeet Singh, Stephen Tu, Noah Brown, Peng Xu, Leila Takayama, Fei Xia, Jake Varley, et~al.
\newblock Robots that ask for help: Uncertainty alignment for large language model planners.
\newblock \emph{arXiv preprint arXiv:2307.01928}, 2023.

\bibitem[Richards(2023)]{autogpt}
Toran~Bruce Richards.
\newblock {AutoGPT}, 2023.
\newblock URL \url{https://github.com/Significant-Gravitas/AutoGPT}.

\bibitem[Rozi{\`e}re et~al.(2023)Rozi{\`e}re, Gehring, Gloeckle, Sootla, Gat, Tan, Adi, Liu, Remez, Rapin, Kozhevnikov, Evtimov, Bitton, Bhatt, Ferrer, Grattafiori, Xiong, D'efossez, Copet, Azhar, Touvron, Martin, Usunier, Scialom, and Synnaeve]{Rozire2023CodeLO}
Baptiste Rozi{\`e}re, Jonas Gehring, Fabian Gloeckle, Sten Sootla, Itai Gat, Xiaoqing Tan, Yossi Adi, Jingyu Liu, Tal Remez, J{\'e}r{\'e}my Rapin, Artyom Kozhevnikov, I.~Evtimov, Joanna Bitton, Manish~P Bhatt, Cristian~Cant{\'o}n Ferrer, Aaron Grattafiori, Wenhan Xiong, Alexandre D'efossez, Jade Copet, Faisal Azhar, Hugo Touvron, Louis Martin, Nicolas Usunier, Thomas Scialom, and Gabriel Synnaeve.
\newblock Code llama: Open foundation models for code.
\newblock \emph{ArXiv}, abs/2308.12950, 2023.

\bibitem[Rozi{\`{e}}re et~al.(2023)Rozi{\`{e}}re, Gehring, Gloeckle, Sootla, Gat, Tan, Adi, Liu, Remez, Rapin, Kozhevnikov, Evtimov, Bitton, Bhatt, Canton{-}Ferrer, Grattafiori, Xiong, D{\'{e}}fossez, Copet, Azhar, Touvron, Martin, Usunier, Scialom, and Synnaeve]{DBLP:journals/corr/abs-2308-12950}
Baptiste Rozi{\`{e}}re, Jonas Gehring, Fabian Gloeckle, Sten Sootla, Itai Gat, Xiaoqing~Ellen Tan, Yossi Adi, Jingyu Liu, Tal Remez, J{\'{e}}r{\'{e}}my Rapin, Artyom Kozhevnikov, Ivan Evtimov, Joanna Bitton, Manish Bhatt, Cristian Canton{-}Ferrer, Aaron Grattafiori, Wenhan Xiong, Alexandre D{\'{e}}fossez, Jade Copet, Faisal Azhar, Hugo Touvron, Louis Martin, Nicolas Usunier, Thomas Scialom, and Gabriel Synnaeve.
\newblock Code llama: Open foundation models for code.
\newblock \emph{CoRR}, abs/2308.12950, 2023.
\newblock \doi{10.48550/arXiv.2308.12950}.
\newblock URL \url{https://doi.org/10.48550/arXiv.2308.12950}.

\bibitem[Sanh et~al.(2022)Sanh, Webson, Raffel, Bach, Sutawika, Alyafeai, Chaffin, Stiegler, Raja, Dey, Bari, Xu, Thakker, Sharma, Szczechla, Kim, Chhablani, Nayak, Datta, Chang, Jiang, Wang, Manica, Shen, Yong, Pandey, Bawden, Wang, Neeraj, Rozen, Sharma, Santilli, F{\'{e}}vry, Fries, Teehan, Scao, Biderman, Gao, Wolf, and Rush]{DBLP:conf/iclr/SanhWRBSACSRDBX22}
Victor Sanh, Albert Webson, Colin Raffel, Stephen~H. Bach, Lintang Sutawika, Zaid Alyafeai, Antoine Chaffin, Arnaud Stiegler, Arun Raja, Manan Dey, M~Saiful Bari, Canwen Xu, Urmish Thakker, Shanya~Sharma Sharma, Eliza Szczechla, Taewoon Kim, Gunjan Chhablani, Nihal~V. Nayak, Debajyoti Datta, Jonathan Chang, Mike~Tian{-}Jian Jiang, Han Wang, Matteo Manica, Sheng Shen, Zheng~Xin Yong, Harshit Pandey, Rachel Bawden, Thomas Wang, Trishala Neeraj, Jos Rozen, Abheesht Sharma, Andrea Santilli, Thibault F{\'{e}}vry, Jason~Alan Fries, Ryan Teehan, Teven~Le Scao, Stella Biderman, Leo Gao, Thomas Wolf, and Alexander~M. Rush.
\newblock Multitask prompted training enables zero-shot task generalization.
\newblock In \emph{The Tenth International Conference on Learning Representations, {ICLR} 2022, Virtual Event, April 25-29, 2022}. OpenReview.net, 2022.
\newblock URL \url{https://openreview.net/forum?id=9Vrb9D0WI4}.

\bibitem[Schick et~al.(2023)Schick, Dwivedi-Yu, Dess{\`\i}, Raileanu, Lomeli, Zettlemoyer, Cancedda, and Scialom]{schick2023toolformer}
Timo Schick, Jane Dwivedi-Yu, Roberto Dess{\`\i}, Roberta Raileanu, Maria Lomeli, Luke Zettlemoyer, Nicola Cancedda, and Thomas Scialom.
\newblock Toolformer: Language models can teach themselves to use tools.
\newblock \emph{arXiv preprint arXiv:2302.04761}, 2023.

\bibitem[Shinn et~al.(2023)Shinn, Cassano, Labash, Gopinath, Narasimhan, and Yao]{shinn2023reflexion}
Noah Shinn, Federico Cassano, Beck Labash, Ashwin Gopinath, Karthik Narasimhan, and Shunyu Yao.
\newblock Reflexion: Language agents with verbal reinforcement learning.
\newblock \emph{arXiv preprint arXiv:2303.11366}, 2023.

\bibitem[Su et~al.(2023)Su, Lan, Li, Xu, Wang, and Cai]{DBLP:journals/corr/abs-2305-16355}
Yixuan Su, Tian Lan, Huayang Li, Jialu Xu, Yan Wang, and Deng Cai.
\newblock Pandagpt: One model to instruction-follow them all.
\newblock \emph{CoRR}, abs/2305.16355, 2023.
\newblock \doi{10.48550/arXiv.2305.16355}.
\newblock URL \url{https://doi.org/10.48550/arXiv.2305.16355}.

\bibitem[Sumers et~al.(2023)Sumers, Yao, Narasimhan, and Griffiths]{sumers2023cognitive}
Theodore Sumers, Shunyu Yao, Karthik Narasimhan, and Thomas~L. Griffiths.
\newblock Cognitive architectures for language agents, 2023.

\bibitem[Taori et~al.(2023)Taori, Gulrajani, Zhang, Dubois, Li, Guestrin, Liang, and Hashimoto]{alpaca}
Rohan Taori, Ishaan Gulrajani, Tianyi Zhang, Yann Dubois, Xuechen Li, Carlos Guestrin, Percy Liang, and Tatsunori~B. Hashimoto.
\newblock Stanford alpaca: An instruction-following llama model.
\newblock \url{https://github.com/tatsu-lab/stanford_alpaca}, 2023.

\bibitem[Touvron et~al.(2023{\natexlab{a}})Touvron, Lavril, Izacard, Martinet, Lachaux, Lacroix, Rozi{\`e}re, Goyal, Hambro, Azhar, et~al.]{touvron2023llama}
Hugo Touvron, Thibaut Lavril, Gautier Izacard, Xavier Martinet, Marie-Anne Lachaux, Timoth{\'e}e Lacroix, Baptiste Rozi{\`e}re, Naman Goyal, Eric Hambro, Faisal Azhar, et~al.
\newblock Llama: Open and efficient foundation language models.
\newblock \emph{arXiv preprint arXiv:2302.13971}, 2023{\natexlab{a}}.

\bibitem[Touvron et~al.(2023{\natexlab{b}})Touvron, Martin, Stone, Albert, Almahairi, Babaei, Bashlykov, Batra, Bhargava, Bhosale, Bikel, Blecher, Canton{-}Ferrer, Chen, Cucurull, Esiobu, Fernandes, Fu, Fu, Fuller, Gao, Goswami, Goyal, Hartshorn, Hosseini, Hou, Inan, Kardas, Kerkez, Khabsa, Kloumann, Korenev, Koura, Lachaux, Lavril, Lee, Liskovich, Lu, Mao, Martinet, Mihaylov, Mishra, Molybog, Nie, Poulton, Reizenstein, Rungta, Saladi, Schelten, Silva, Smith, Subramanian, Tan, Tang, Taylor, Williams, Kuan, Xu, Yan, Zarov, Zhang, Fan, Kambadur, Narang, Rodriguez, Stojnic, Edunov, and Scialom]{DBLP:journals/corr/abs-2307-09288}
Hugo Touvron, Louis Martin, Kevin Stone, Peter Albert, Amjad Almahairi, Yasmine Babaei, Nikolay Bashlykov, Soumya Batra, Prajjwal Bhargava, Shruti Bhosale, Dan Bikel, Lukas Blecher, Cristian Canton{-}Ferrer, Moya Chen, Guillem Cucurull, David Esiobu, Jude Fernandes, Jeremy Fu, Wenyin Fu, Brian Fuller, Cynthia Gao, Vedanuj Goswami, Naman Goyal, Anthony Hartshorn, Saghar Hosseini, Rui Hou, Hakan Inan, Marcin Kardas, Viktor Kerkez, Madian Khabsa, Isabel Kloumann, Artem Korenev, Punit~Singh Koura, Marie{-}Anne Lachaux, Thibaut Lavril, Jenya Lee, Diana Liskovich, Yinghai Lu, Yuning Mao, Xavier Martinet, Todor Mihaylov, Pushkar Mishra, Igor Molybog, Yixin Nie, Andrew Poulton, Jeremy Reizenstein, Rashi Rungta, Kalyan Saladi, Alan Schelten, Ruan Silva, Eric~Michael Smith, Ranjan Subramanian, Xiaoqing~Ellen Tan, Binh Tang, Ross Taylor, Adina Williams, Jian~Xiang Kuan, Puxin Xu, Zheng Yan, Iliyan Zarov, Yuchen Zhang, Angela Fan, Melanie Kambadur, Sharan Narang, Aur{\'{e}}lien Rodriguez, Robert Stojnic, Sergey Edunov,
  and Thomas Scialom.
\newblock Llama 2: Open foundation and fine-tuned chat models.
\newblock \emph{CoRR}, abs/2307.09288, 2023{\natexlab{b}}.
\newblock \doi{10.48550/arXiv.2307.09288}.
\newblock URL \url{https://doi.org/10.48550/arXiv.2307.09288}.

\bibitem[Vemprala et~al.(2023)Vemprala, Bonatti, Bucker, and Kapoor]{DBLP:journals/corr/abs-2306-17582}
Sai Vemprala, Rogerio Bonatti, Arthur Bucker, and Ashish Kapoor.
\newblock Chatgpt for robotics: Design principles and model abilities.
\newblock \emph{CoRR}, abs/2306.17582, 2023.
\newblock \doi{10.48550/arXiv.2306.17582}.
\newblock URL \url{https://doi.org/10.48550/arXiv.2306.17582}.

\bibitem[Wang et~al.(2023{\natexlab{a}})Wang, Xie, Jiang, Mandlekar, Xiao, Zhu, Fan, and Anandkumar]{wang2023voyager}
Guanzhi Wang, Yuqi Xie, Yunfan Jiang, Ajay Mandlekar, Chaowei Xiao, Yuke Zhu, Linxi Fan, and Anima Anandkumar.
\newblock Voyager: An open-ended embodied agent with large language models.
\newblock \emph{arXiv preprint arXiv:2305.16291}, 2023{\natexlab{a}}.

\bibitem[Wang et~al.(2023{\natexlab{b}})Wang, Ma, Feng, Zhang, Yang, Zhang, Chen, Tang, Chen, Lin, Zhao, Wei, and Wen]{wang2023survey}
Lei Wang, Chen Ma, Xueyang Feng, Zeyu Zhang, Hao Yang, Jingsen Zhang, Zhiyuan Chen, Jiakai Tang, Xu~Chen, Yankai Lin, Wayne~Xin Zhao, Zhewei Wei, and Ji-Rong Wen.
\newblock A survey on large language model based autonomous agents, 2023{\natexlab{b}}.

\bibitem[Wang et~al.(2023{\natexlab{c}})Wang, Zhou, Zu, Xia, Chen, Zhang, Zheng, Ye, Zhang, Gui, Kang, Yang, Li, and Du]{DBLP:journals/corr/abs-2304-08085}
Xiao Wang, Weikang Zhou, Can Zu, Han Xia, Tianze Chen, Yuansen Zhang, Rui Zheng, Junjie Ye, Qi~Zhang, Tao Gui, Jihua Kang, Jingsheng Yang, Siyuan Li, and Chunsai Du.
\newblock Instructuie: Multi-task instruction tuning for unified information extraction.
\newblock \emph{CoRR}, abs/2304.08085, 2023{\natexlab{c}}.
\newblock \doi{10.48550/arXiv.2304.08085}.
\newblock URL \url{https://doi.org/10.48550/arXiv.2304.08085}.

\bibitem[Wang et~al.(2023{\natexlab{d}})Wang, Kordi, Mishra, Liu, Smith, Khashabi, and Hajishirzi]{DBLP:conf/acl/WangKMLSKH23}
Yizhong Wang, Yeganeh Kordi, Swaroop Mishra, Alisa Liu, Noah~A. Smith, Daniel Khashabi, and Hannaneh Hajishirzi.
\newblock Self-instruct: Aligning language models with self-generated instructions.
\newblock In Anna Rogers, Jordan~L. Boyd{-}Graber, and Naoaki Okazaki (eds.), \emph{Proceedings of the 61st Annual Meeting of the Association for Computational Linguistics (Volume 1: Long Papers), {ACL} 2023, Toronto, Canada, July 9-14, 2023}, pp.\  13484--13508. Association for Computational Linguistics, 2023{\natexlab{d}}.
\newblock \doi{10.18653/v1/2023.acl-long.754}.
\newblock URL \url{https://doi.org/10.18653/v1/2023.acl-long.754}.

\bibitem[Wei et~al.(2022{\natexlab{a}})Wei, Bosma, Zhao, Guu, Yu, Lester, Du, Dai, and Le]{wei2022finetuned}
Jason Wei, Maarten Bosma, Vincent~Y. Zhao, Kelvin Guu, Adams~Wei Yu, Brian Lester, Nan Du, Andrew~M. Dai, and Quoc~V. Le.
\newblock Finetuned language models are zero-shot learners, 2022{\natexlab{a}}.

\bibitem[Wei et~al.(2022{\natexlab{b}})Wei, Wang, Schuurmans, Bosma, Chi, Le, and Zhou]{wei2022chain}
Jason Wei, Xuezhi Wang, Dale Schuurmans, Maarten Bosma, Ed~Chi, Quoc Le, and Denny Zhou.
\newblock Chain of thought prompting elicits reasoning in large language models.
\newblock \emph{arXiv preprint arXiv:2201.11903}, 2022{\natexlab{b}}.

\bibitem[Weng(2023)]{weng2023prompt}
Lilian Weng.
\newblock Llm-powered autonomous agents.
\newblock \emph{lilianweng.github.io}, Jun 2023.
\newblock URL \url{https://lilianweng.github.io/posts/2023-06-23-agent/}.

\bibitem[Willison(2023)]{willison2023injection}
Simon Willison.
\newblock Prompt injection: What’s the worst that can happen?
\newblock \emph{https://simonwillison.net}, 2023.
\newblock URL \url{https://simonwillison.net/2023/Apr/14/worst-that-can-happen/}.

\bibitem[Xi et~al.(2023)Xi, Chen, Guo, He, Ding, Hong, Zhang, Wang, Jin, Zhou, et~al.]{xi2023rise}
Zhiheng Xi, Wenxiang Chen, Xin Guo, Wei He, Yiwen Ding, Boyang Hong, Ming Zhang, Junzhe Wang, Senjie Jin, Enyu Zhou, et~al.
\newblock The rise and potential of large language model based agents: A survey.
\newblock \emph{arXiv preprint arXiv:2309.07864}, 2023.

\bibitem[Xu et~al.(2023)Xu, Sun, Zheng, Geng, Zhao, Feng, Tao, and Jiang]{DBLP:journals/corr/abs-2304-12244}
Can Xu, Qingfeng Sun, Kai Zheng, Xiubo Geng, Pu~Zhao, Jiazhan Feng, Chongyang Tao, and Daxin Jiang.
\newblock Wizardlm: Empowering large language models to follow complex instructions.
\newblock \emph{CoRR}, abs/2304.12244, 2023.
\newblock \doi{10.48550/arXiv.2304.12244}.
\newblock URL \url{https://doi.org/10.48550/arXiv.2304.12244}.

\bibitem[Yang et~al.(2018)Yang, Qi, Zhang, Bengio, Cohen, Salakhutdinov, and Manning]{yang-etal-2018-hotpotqa}
Zhilin Yang, Peng Qi, Saizheng Zhang, Yoshua Bengio, William Cohen, Ruslan Salakhutdinov, and Christopher~D. Manning.
\newblock {H}otpot{QA}: A dataset for diverse, explainable multi-hop question answering.
\newblock In \emph{Proceedings of the 2018 Conference on Empirical Methods in Natural Language Processing}, pp.\  2369--2380, Brussels, Belgium, October-November 2018. Association for Computational Linguistics.
\newblock \doi{10.18653/v1/D18-1259}.
\newblock URL \url{https://aclanthology.org/D18-1259}.

\bibitem[Yao et~al.(2022{\natexlab{a}})Yao, Chen, Yang, and Narasimhan]{yao2022webshop}
Shunyu Yao, Howard Chen, John Yang, and Karthik Narasimhan.
\newblock Webshop: Towards scalable real-world web interaction with grounded language agents.
\newblock \emph{Advances in Neural Information Processing Systems}, 35:\penalty0 20744--20757, 2022{\natexlab{a}}.

\bibitem[Yao et~al.(2022{\natexlab{b}})Yao, Zhao, Yu, Du, Shafran, Narasimhan, and Cao]{yao2022react}
Shunyu Yao, Jeffrey Zhao, Dian Yu, Nan Du, Izhak Shafran, Karthik Narasimhan, and Yuan Cao.
\newblock React: Synergizing reasoning and acting in language models.
\newblock \emph{arXiv preprint arXiv:2210.03629}, 2022{\natexlab{b}}.

\bibitem[Yao et~al.(2023)Yao, Yu, Zhao, Shafran, Griffiths, Cao, and Narasimhan]{yao2023tree}
Shunyu Yao, Dian Yu, Jeffrey Zhao, Izhak Shafran, Thomas~L Griffiths, Yuan Cao, and Karthik Narasimhan.
\newblock Tree of thoughts: Deliberate problem solving with large language models.
\newblock \emph{arXiv preprint arXiv:2305.10601}, 2023.

\bibitem[Zelikman et~al.(2022)Zelikman, Wu, Mu, and Goodman]{zelikman2022star}
Eric Zelikman, Yuhuai Wu, Jesse Mu, and Noah Goodman.
\newblock {STaR: Bootstrapping reasoning with reasoning}.
\newblock \emph{Advances in Neural Information Processing Systems}, 35:\penalty0 15476--15488, 2022.

\bibitem[Zhang et~al.(2023{\natexlab{a}})Zhang, Li, and Bing]{DBLP:journals/corr/abs-2306-02858}
Hang Zhang, Xin Li, and Lidong Bing.
\newblock Video-llama: An instruction-tuned audio-visual language model for video understanding.
\newblock \emph{CoRR}, abs/2306.02858, 2023{\natexlab{a}}.
\newblock \doi{10.48550/arXiv.2306.02858}.
\newblock URL \url{https://doi.org/10.48550/arXiv.2306.02858}.

\bibitem[Zhang et~al.(2023{\natexlab{b}})Zhang, Dong, Li, Zhang, Sun, Wang, Li, Hu, Zhang, Wu, and Wang]{DBLP:journals/corr/abs-2308-10792}
Shengyu Zhang, Linfeng Dong, Xiaoya Li, Sen Zhang, Xiaofei Sun, Shuhe Wang, Jiwei Li, Runyi Hu, Tianwei Zhang, Fei Wu, and Guoyin Wang.
\newblock Instruction tuning for large language models: {A} survey.
\newblock \emph{CoRR}, abs/2308.10792, 2023{\natexlab{b}}.
\newblock \doi{10.48550/arXiv.2308.10792}.
\newblock URL \url{https://doi.org/10.48550/arXiv.2308.10792}.

\bibitem[Zhang et~al.(2023{\natexlab{c}})Zhang, Zhang, Gu, Zhou, Lipka, Yang, and Sun]{DBLP:journals/corr/abs-2306-17107}
Yanzhe Zhang, Ruiyi Zhang, Jiuxiang Gu, Yufan Zhou, Nedim Lipka, Diyi Yang, and Tong Sun.
\newblock Llavar: Enhanced visual instruction tuning for text-rich image understanding.
\newblock \emph{CoRR}, abs/2306.17107, 2023{\natexlab{c}}.
\newblock \doi{10.48550/arXiv.2306.17107}.
\newblock URL \url{https://doi.org/10.48550/arXiv.2306.17107}.

\bibitem[Zhang et~al.(2020)Zhang, Sun, Galley, Chen, Brockett, Gao, Gao, Liu, and Dolan]{zhang2020dialogpt}
Yizhe Zhang, Siqi Sun, Michel Galley, Yen-Chun Chen, Chris Brockett, Xiang Gao, Jianfeng Gao, Jingjing Liu, and William~B Dolan.
\newblock Dialogpt: Large-scale generative pre-training for conversational response generation.
\newblock In \emph{Proceedings of the 58th Annual Meeting of the Association for Computational Linguistics: System Demonstrations}, pp.\  270--278, 2020.

\bibitem[Zhou et~al.(2023{\natexlab{a}})Zhou, Liu, Xu, Iyer, Sun, Mao, Ma, Efrat, Yu, Yu, Zhang, Ghosh, Lewis, Zettlemoyer, and Levy]{DBLP:journals/corr/abs-2305-11206}
Chunting Zhou, Pengfei Liu, Puxin Xu, Srini Iyer, Jiao Sun, Yuning Mao, Xuezhe Ma, Avia Efrat, Ping Yu, Lili Yu, Susan Zhang, Gargi Ghosh, Mike Lewis, Luke Zettlemoyer, and Omer Levy.
\newblock {LIMA:} less is more for alignment.
\newblock \emph{CoRR}, abs/2305.11206, 2023{\natexlab{a}}.
\newblock \doi{10.48550/arXiv.2305.11206}.
\newblock URL \url{https://doi.org/10.48550/arXiv.2305.11206}.

\bibitem[Zhou et~al.(2023{\natexlab{b}})Zhou, Xu, Zhu, Zhou, Lo, Sridhar, Cheng, Bisk, Fried, Alon, et~al.]{zhou2023webarena}
Shuyan Zhou, Frank~F Xu, Hao Zhu, Xuhui Zhou, Robert Lo, Abishek Sridhar, Xianyi Cheng, Yonatan Bisk, Daniel Fried, Uri Alon, et~al.
\newblock {WebArena: A Realistic Web Environment for Building Autonomous Agents}.
\newblock \emph{arXiv preprint arXiv:2307.13854}, 2023{\natexlab{b}}.

\end{thebibliography}
\bibliographystyle{iclr2024_conference}

\newpage
\appendix
\section{Additional Results}
\subsection{Standard Error of EM Scores}
\label{app:std}

\begin{table}[h]
\centering
\caption{Prompting results with Standard Errors (\(\sigma_M\)).}
\begin{tabular}{lccc}
\toprule
                    & Prompt        & EM            & \(\sigma_M\) \\ \hline
\multirow{3}{*}{GPT-4} & IO   & 37.2          & 2.16 \\
                       & \cott{}   & \textbf{45.0} & 2.22 \\
                       & \react{} & 42.0          & 2.21 \\ \hline
\multirow{3}{*}{GPT-3.5} & IO   & 22.4          & 1.86 \\
                         & \cott{}   & 28.0          & 2.01 \\
                         & \react{} & \textbf{31.4} & 2.08 \\ 
\bottomrule
\end{tabular}
\label{tab:hotpot_baseline_std}
\end{table}

\begin{table}[h]
\centering
\caption{Prompting vs. fine-tuning, with absolute/relative increases and Standard Errors ($\sigma_M$).}
\begin{tabular}{lcccccc}
\toprule
            & \react{}        & \method{} & abs./rel. diff & \(\sigma_{\text{ReAct}}\) & \(\sigma_{\text{FireAct}}\) \\ \hline
Llama-2-7B   & 14.8                     & 26.2        & +11.4 / \textbf{77\%} & 1.59 & 1.97 \\
Llama-2-13B  & \multicolumn{1}{c}{21.2} & 34.4        & \textbf{+13.1} / 62\% & 1.83 & 2.12 \\
CodeLlama-7B & 17.4                     & 27.8        & +10.4 / 60\% & 1.69 & 2.00 \\
CodeLlama-13B& 20.8                     & 29.0        & +8.2 / 39\% & 1.82 & 2.03 \\
CodeLlama-34B& \multicolumn{1}{c}{22.2} & 27.8        & +5.6 / 25\% & 1.86 & 2.00 \\
GPT-3.5      & \textbf{31.4}             & \textbf{39.2} & +7.8 / 25\% & 2.08 & 2.18 \\
\bottomrule
\end{tabular}
\label{tab:few_vs_ft_std}
\end{table}

\begin{table}[h]
\centering
\caption{Multi-task results with GPT-3.5 and Standard Errors (\(\sigma_M\)).}
\begin{tabular}{@{}lcccccccc@{}}
\toprule
                   & \multicolumn{4}{c}{\textbf{HotpotQA}} & \multicolumn{4}{c}{\textbf{StrategyQA}} \\ \midrule
                   & EM & \(\sigma_M\) & EM & \(\sigma_M\) & EM & \(\sigma_M\) & EM & \(\sigma_M\) \\
\textbf{Prompting}  &      &            &      &            &      &            &      & \\
IO                 & 22.4 & 1.86     & 48.0 & 2.23     & 7.2  & 1.16     & 68.6 & 2.08 \\
\cott{}            & 28.0 & 2.01     & 49.0 & 2.24     & 41.6 & 2.20     & 50.8 & 2.24 \\
\react{}           & 31.4 & 2.08     & 61.0 & 2.18     & 40.8 & 2.20     & 58.6 & 2.20 \\ \midrule
\textbf{Fine-tuning} &      &            &      &            &      &            &      & \\
HotpotQA           & 39.2 & 2.18     & -    & -          & 44.0 & 2.22     & -    & - \\
Multi-task         & 39.2 & 2.18     & 55.5 & 2.19     & 43.2 & 2.20     & 63.2 & 2.16 \\
+ \cott{}          & \textbf{39.6} & 2.19 & \textbf{72.9} & 1.99 & \textbf{50.4} & 2.24 & 65.8 & 2.12 \\ \bottomrule
\end{tabular}
\label{tab:multi_task_std}
\end{table}

\subsection{Robustness Analysis}
\label{app:robustness}
\begin{table}[t]
\centering
\begin{tabular}{ll|ccc}
\toprule
                         &                           & Normal                   & None                     & Random                   \\
\multicolumn{2}{l|}{\react{}}                           & \multicolumn{1}{c}{31.4} & \multicolumn{1}{c}{20.8} & \multicolumn{1}{c}{22.6} \\ \hline
\multirow{4}{*}{\method{}}& Single Task Single Method & 39.2                     & 33.6                     & 37.2                     \\
                         & Multi-task, Single Method & 39.2                     & 34.8                     & 36.2                     \\
                         & Single Task, Multi-method & 41.0                       & 36.4                     & 38.4                     \\
                         & Multi-task, Multi-method  & 39.6                     & 35.2                     & 37.0                       \\ \bottomrule
\end{tabular}
\caption{Comparison of results for different tasks and methods.}
\end{table}

\subsection{Data Mix}

\begin{table}[ht]
\centering
\begin{tabular}{lllllccc}
\toprule
\multicolumn{5}{l}{\multirow{2}{*}{\textit{Multi-methods Language Agent Finetuning}}}     &      & \multicolumn{2}{c}{Number of Turns} \\
\multicolumn{5}{l}{}                                                                      & EM   & $\mu$           & $\sigma$          \\ \hline
Single Agent Method   & \multicolumn{4}{l}{IO (Prompt)}                                   & 22.4 & 0               & 0                 \\
                      & \multicolumn{4}{l}{\cott{} (Prompt)}                                  & 28.0 & 0.8             & 0.4               \\
                      & \multicolumn{4}{l}{\react{} (Finetune)}                              & 39.4 & 3.2             & 1.4               \\
                      & \multicolumn{4}{l}{\reflexion{} (Finetune)}                         & 39.8 & 3.5             & 2.6               \\ %
\multicolumn{8}{l}{}                                                                                                                   \\ \toprule
\multicolumn{5}{l}{\multirow{2}{*}{\textit{Chosen Agent Method}}}                         &      & \multicolumn{2}{c}{Number of Turns} \\
\multicolumn{5}{l}{}                                                                      & EM   & $\mu$           & $\sigma$          \\ \hline
Practical             & \multicolumn{4}{l}{Randomly choose one}                           & 32.4 & -               & -                 \\
                      & \multicolumn{4}{l}{Best single method}                            & 39.8 & 3.5             & 2.6               \\ 
Theoretically optimal & \multicolumn{4}{l}{Always choose the best one}                    & 52.0 & 1.3             & 1.1               \\ %
\multicolumn{8}{l}{}                                                                                                                   \\ \toprule
\multirow{2}{*}{\textit{Mix agent methods \method{}}} &              &              &              & \multicolumn{1}{c}{} &      & \multicolumn{2}{c}{Number of Turns} \\
                                                    & \react{}        & IO           & \cott{}          & \reflexion{}           & EM   & $\mu$           & $\sigma$          \\ \hline
                      & $\checkmark$ & $\checkmark$ & $\checkmark$ & $\checkmark$                  & 39.4 & 2.4             & 1.8               \\
                      & $\checkmark$ & $\checkmark$ & $\checkmark$          & $\times$                   & 41.0 & 2.6             & 1.6               \\
                      & $\checkmark$ & $\checkmark$ & $\times$           & $\checkmark$                  & 41.2 & 3.3             & 1.6               \\
                      & $\checkmark$ & $\checkmark$ & $\times$           & $\times$                   & 40.2 & 3.3             & 1.5               \\
                      & $\checkmark$ & $\times$ & $\checkmark$          & $\checkmark$                  & 40.0 & 3.0             & 4.8               \\
                      & $\checkmark$ & $\times$ & $\checkmark$          & $\times$                   & 41.0 & 2.7             & 1.7               \\
                      & $\checkmark$ & $\times$ & $\times$           & $\checkmark$                  & 38.8 & 3.8             & 2.8               \\\bottomrule
\end{tabular}
\caption{Mixed agent methods investigation on HotpotQA}
\label{tab:data_mix_study_appendix}
\end{table}

The Table \ref{tab:data_mix_study_appendix} illustrates the lower and upper performance limits of the model. The lower boundary is determined by randomly selecting one of the agent methods, while the upper boundary is established by always selecting the best agent method when a question is asked. The lower and upper bounds yielded 32.4 and 52.0 (EM), respectively. The goal of the language agent fine-tuning with mixed agent methods is to reach the theoretical optimum, which is impractical but optimal. The results of the data mix demonstrate that the performance of all data mix falls within the range of the mean and the best baselines, indicating a consistent improvement. Notably, the inclusion of the \cott{} method is especially advantageous, resulting in a significant improvement in performance. Despite the relatively poor performance of \cott{} individually (28.0 EM as suggested in Table \ref{tab:hotpot_baseline}), their combined data contribute positively to the fine-tuning process. Table \ref{tab:data_mix_study_appendix} also summarizes our research into the effects of mixed agent methods applied to fine-tuned language agents. The results are varied, with the EM performance of the agent, when various mixed methods are used, ranging from 38.8 to 41.0.
\subsection{Turn Distribution}
\begin{figure}[ht]
   \begin{minipage}{0.55\linewidth}
        \centering
        \includegraphics[width=\linewidth]{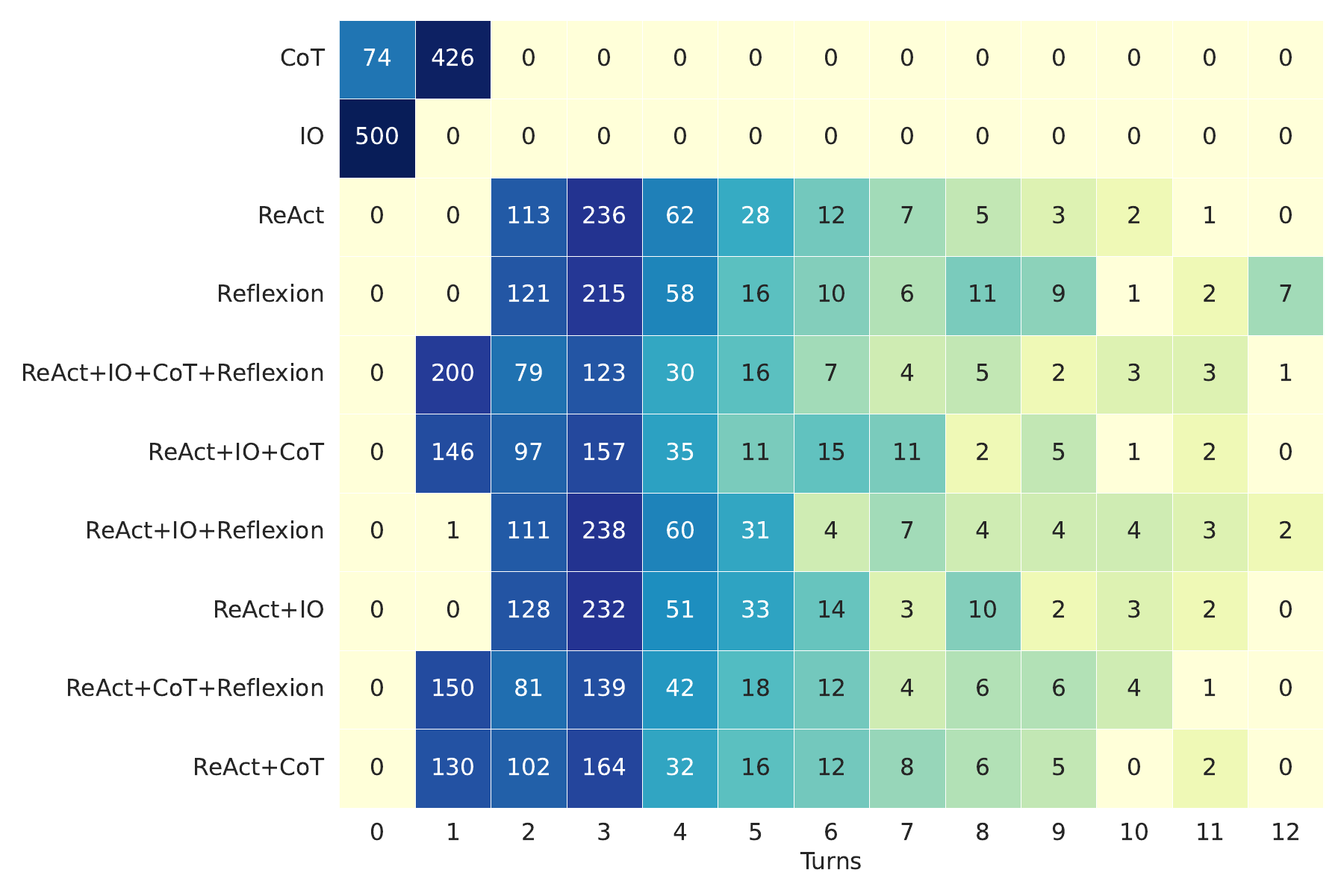}
    \end{minipage}
    \begin{minipage}{0.45\linewidth}
        \centering
        \includegraphics[width=\linewidth]{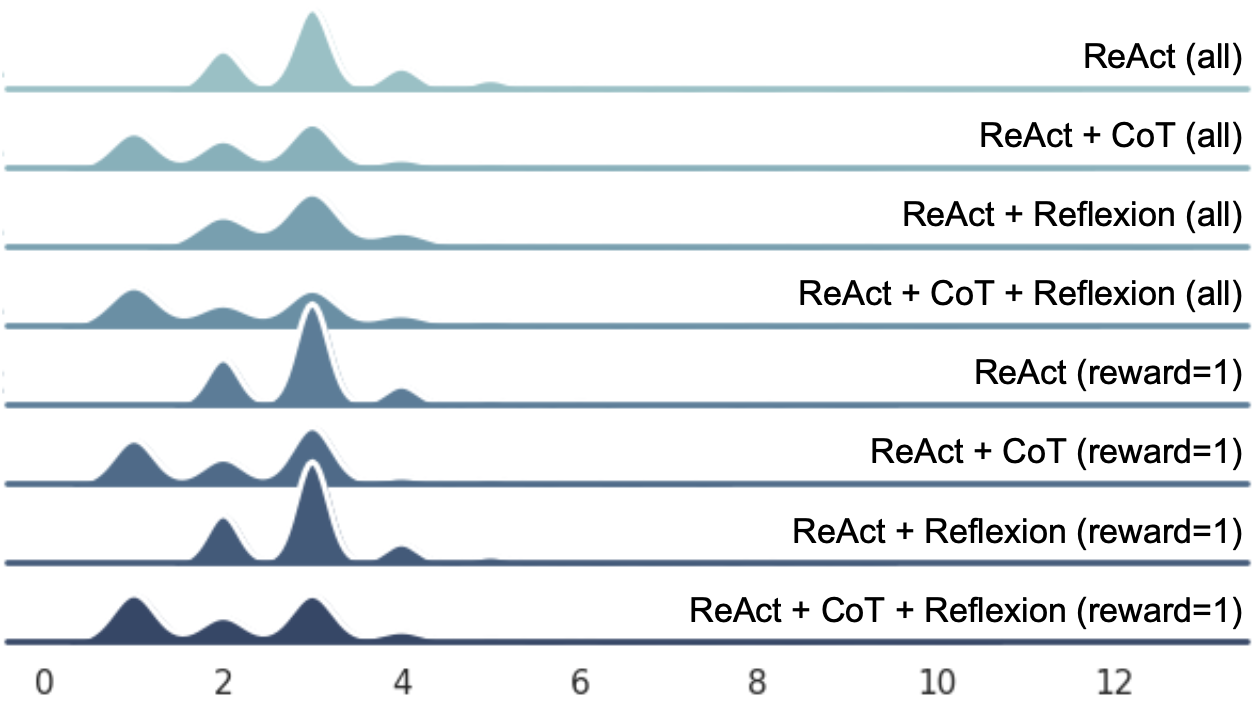}
    \end{minipage}%
    \caption{Turn distribution}
    \label{fig:turn_dist}
\end{figure}

The graph in Figure \ref{fig:turn_dist} demonstrates that combining data changes the distribution to resemble that of the training data. This was especially noticeable when contrasting \react{}+\cott{} (2.7 turns on average, 41.0 EM) and \react{}+\cott{} (3.8 turns on average, 38.8 EM).

\subsection{LoRA vs. Full Fine-tuning}
\label{app:lora_full}

\begin{table}[ht]
    \centering
\begin{tabular}{lcc}
\toprule
                   & LoRA                     & Full                     \\ \hline
$EM_{\text{single}}$ & 26.2                     & 30.2                     \\
$EM_{\text{multi}}$  & 27.0                     & 30.0                     \\
$M_{\text{size}}$    & 38MB & 14GB \\ \bottomrule
\end{tabular}
\caption{LoRA vs. Full fine-tuning on single-task setting and multi-task setting.}
\label{tab:lora_vs_full}
\end{table}
Table \ref{tab:lora_vs_full} demonstrates the EM score of LLama-2-7B fine-tuned on 500 HotpotQA trajectories (single-task setting) and mixed all trajectories (multi-task setting) with LoRA and full weights respectively.

\subsection{CodeLLama vs. Llama Tokenizer}
\label{app:code_tokenizer}
\begin{table}[h]
\centering
\begin{tabular}{lccc}
\toprule
              & Llama Tokenizer & CodeLLama Tokenizer & abs./rel. diff \\ \hline
CodeLlama-7B  & 26.6            & 27.8                & +1.2 / 4.5\%   \\
CodeLlama-13B & 27.4            & 29.0                & +1.6 / 5.8\%   \\ \bottomrule
\end{tabular}
\caption{The performance difference of CodeLlama between using Llama tokenizers and CodeLlama tokenizers.}
\label{tab:code_tok}
\end{table}

Table \ref{tab:code_tok} compares the Llama Tokenizer and the CodeLLama Tokenizer in terms of their impact on the performance of the CodeLlama model. EM scores are provided for two variants of the CodeLlama model, CodeLlama-7B and CodeLlama-13B, for each tokenizer on HotpotQA. 
\subsection{World modelling: masking observations}
Table \ref{tab:mask_model} suggests that the incorporation of observation masking generally leads to slight improvements in performance for the CodeLlama-7B model. The most significant improvement is seen in the ReAct + CoT setting. However, the influence of observation masking on the CodeLlama-13B model is not consistent across configurations, with some cases showing improvement and others not. This table provides a detailed overview of the effect of observation masking on the CodeLlama models' performance in HotpotQA tasks, with the performance metric values for each configuration presented for further study. It appears that in our current settings, learning world modelling does not have a consistent effect on performance.
\begin{table}[ht]
\centering
\begin{tabular}{lcc}
\toprule
                        & w/ Observation Mask & w/o Observation Mask \\ \hline
\multicolumn{3}{l}{\textit{CodeLlama-7B}}                            \\
ReAct                   & 26.0                & \textbf{26.2}        \\
ReAct + CoT             & \textbf{24.0}       & 22.2                 \\
ReAct + Reflexion       & \textbf{25.2}       & 24.8                 \\
ReAct + Reflexion + CoT & \textbf{26.0}       & 24.8                 \\ \hline
\multicolumn{3}{l}{\textit{CodeLlama-13B}}                           \\
ReAct                   & \textbf{34.6}       & 34.4                 \\
ReAct + CoT             & 24.8                & \textbf{30.2}        \\
ReAct + Reflexion       & \textbf{30.6}       & \textbf{30.6}        \\
ReAct + Reflexion + CoT & 27.2                & \textbf{31.0}        \\ \bottomrule
\end{tabular}
\caption{The performance difference of fine-tuning Llama-2-7B and Llama-2-13B with or without masking the observations in HotpotQA trajectories.}
\label{tab:mask_model}
\end{table}

\subsection{Training Epochs}
\begin{figure}[t]
    \centering
    \includegraphics[width=0.8\linewidth]{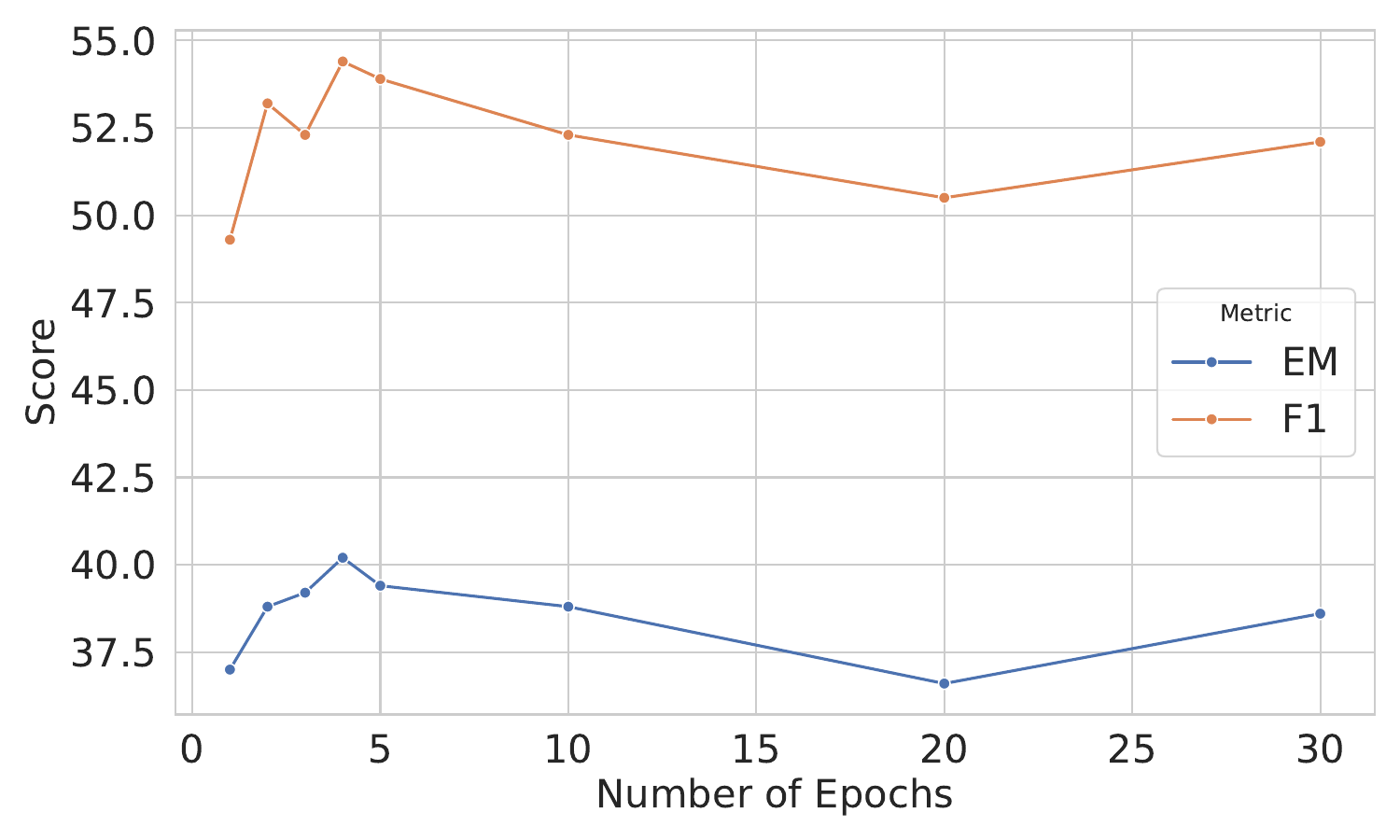}
    \caption{Performance changes of GPT-3.5-Turbo on HotpotQA dataset with increasing fine-tuning epochs}
    \label{fig:hotpotqa_epoches}
\end{figure}

Table \ref{fig:hotpotqa_epoches} presents the performance changes of the GPT-3.5 model on the HotpotQA dataset as the number of fine-tuning epochs varies. As the number of fine-tuning epochs increases from 1 to 4, both the EM and F1 scores generally improve, indicating increased precision in providing exact answers and overall answer quality. However, beyond 4 epochs, while EM scores continue to increase slightly, F1 scores plateau and even dip at times, suggesting a diminishing return on additional fine-tuning.

\subsection{Few-shot Task Generalization of \method{} }
\label{app:few_shot_gen}
\begin{table}[h]
\centering
\begin{tabular}{lccc}
\hline
              & {{StrategyQA}}    & {{Bamboogle}}    & {{MMLU}}    \\ \hline
\multicolumn{4}{l}{\textit{Fine-tuned Llama-2-7B + Few-shot Prompt}} \\
IO            & 44.1              & 0                & 5.4         \\
\cott{}       & 27.0              & 0                & 4.2         \\
\react{}      & 52.0              & 35.2             & 34.0        \\ \hline
\multicolumn{4}{l}{\textit{Vanilla Models + Few-shot ReAct}}       \\
Llama-2-7B    & 59.0              & 11.2             & 33.2        \\
GPT-3.5       & 61.0              & 40.8             & 58.6        \\ \hline
\end{tabular}
\caption{Multi-task results with Llama-2-7B fine-tuned on HotpotQA trajectories.}
\label{tab:adapt_prompt}
\end{table}

\section{Experiment Details}
\label{app:exp_detail}
\subsection{Base Models}
We explored the base models from three representative LLM families:
(1) Generative Pre-trained Transformer (GPT) family that includes GPT-4 \citep{openai2023gpt4} and GPT-3.5, Llama-2 \citep{DBLP:journals/corr/abs-2307-09288}, and CodeLlama \citep{DBLP:journals/corr/abs-2308-12950}. GPT-3.5 is the instructed and reinforcement learning from human feedback (RLHF) version of GPT-3 \citep{brown2020language} with 175B parameters and is capable of understanding and generating human-like text in a wide range of tasks and domains. GPT-4 is widely regarded as one of the state-of-the-art LLMs, and both GPT-3.5 and GPT-4 have the ability to use a sequence of tools in a zero-shot context with the feature of the plugin \footnote{\url{https://openai.com/blog/chatgpt- plugins}}. (2) LLama-2 is a series of open source pre-trained and fine-tuned LLMs ranging from 7B to 70B, which has preliminary evidence of tool use emergence, but its performance has yet to be extensively tested. (3) CodeLlama is a family of LLMs specifically designed for code generation derived from Llama-2 and ranging from 7B to 34B, which gradually specialises and increases the capabilities of Llama-2 models in coding by applying a cascade of training and fine-tuning steps. 

\subsection{Single-method Single-task Setup}
We initially took samples from the HotpotQA Train set and asked GPT-4 models \citep{openai2023gpt4} to generate 500 ReAct trajectories with few-shot examples for agent fine-tuning with human-in-the-loop validation. We also selected 500 examples from the original 7,405 Dev set for evaluation with the exact match (EM) and the F1 score as two metrics. GPT-3.5 was fine-tuned \texttt{GPT-3.5-Turbo-0613} with OpenAI fine-tuning API \footnote{\url{https://platform.openai.com/docs/guides/fine-tuning}}, while Llama-2 and CodeLlama were fine-tuned \texttt{Llama-2-Chat-HF} and \texttt{CodeLlama-Instruct-HF} on A100-80GB, RTX 6000 Ada, RTX 4090 and RTX 3090 GPUs with Low-Rank Adaptation (LoRA) \citep{DBLP:conf/iclr/HuSWALWWC22} with \texttt{int8} quantization. We use OpenAI fine-tuning API~\citep{openai2023finetune} to fine-tune \texttt{GPT-3.5-Turbo-0613} for 3 epochs and Low-Rank Adaptation (LoRA)~\citep{DBLP:conf/iclr/HuSWALWWC22} to fine-tune Llama-2 and CodeLlama families for 30 epochs. We set a hard limit if the \react{} format algorithms did not finish in 11 steps. The learning rate is \texttt{3e-4}, the trainning batch size is \texttt{16}. The evaluation temperature of the Llama and Codellama models is \texttt{0.6}, and the temperature of GPT-3.5 is \texttt{0}. We also evaluate the models also with \texttt{int8} quantization.

\subsection{Computation of Llama and CodeLlama}
Table \ref{tab:gpu_use} demonstrates the computational resources used for fine-tuning and inferring Llama-2 and CodeLlama models.
\begin{table}[h]
\centering
\begin{tabular}{lccc}
\toprule
              & Training GPU                          & Example / Second & Inference GPU                \\ \hline
\multicolumn{3}{l}{\textit{LoRA Fine-tuning}}                            &                              \\
Llama-2-7B    & 1 $\times$ RTX 4090 - 24GB              & 5.4              & 1 $\times$ RTX 4090 - 24GB     \\
Llama-2-13B   & 1 $\times$ RTX 4090 - 24GB     & 2.5              & 1 $\times$ RTX 4090 - 24GB     \\
CodeLlama-7B  & 1 $\times$ RTX 4090 - 24GB              & 5.5              & 1 $\times$ RTX 4090 - 24GB     \\
CodeLlama-13B & 1 $\times$ RTX 4090 - 24GB    & 2.2              & 1 $\times$ RTX 4090 - 24GB     \\
CodeLlama-34B & 1 $\times$ RTX 6000 Ada - 48GB & 0.9              & 1 $\times$ RTX 6000 Ada - 48GB \\ \hline
\multicolumn{3}{l}{\textit{Full Fine-tuning}}                            &                              \\
Llama-2-7B    & 4 $\times$ A100 - 80GB                  & 19.7             & 1 $\times$ RTX 4090 - 24GB     \\ \bottomrule
\end{tabular}
\caption{The computational resources used for training and inference with Llama-2 and CodeLlama models.}
\label{tab:gpu_use}
\end{table}

\newpage
\section{Prompts}
\label{sec:prompts}
\begin{table}[H]
\scriptsize
\begin{minipage}{0.9\linewidth}
Hotpot QA Prompts
\centering
\ttfamily
\begin{tabular}{l p{0.9\linewidth}}
\midrule
\sethlcolor{cyan}\hl{Input-Output(IO)} & \\
Question &  What is the elevation range for the area that the eastern sector of the Colorado orogeny extends into?\\
Answer &  1,800 to 7,000 ft\\
\\
Question &  Musician and satirist Allie Goertz wrote a song about the "The Simpsons" character Milhouse, who Matt Groening named after who?\\
Answer &  Richard Nixon\\
\\
Question &  Which documentary is about Finnish rock groups, Adam Clayton Powell or The Saimaa Gesture?\\
Answer &  The Saimaa Gesture\\
\\
Question &  What profession does Nicholas Ray and Elia Kazan have in common?\\
Answer &  director, screenwriter, actor\\
\\
Question &  Which magazine was started first Arthur's Magazine or First for Women?\\
Answer &  Arthur's Magazine\\
\\
Question &  Were Pavel Urysohn and Leonid Levin known for the same type of work?\\
Answer &  Yes\\

\midrule
\\
\sethlcolor{yellow}\hl{Chain-of-Thoughts (CoT)} & \\
Question &  What is the elevation range for the area that the eastern sector of the Colorado orogeny extends into?\\
Thought &  The eastern sector of Colorado orogeny extends into the High Plains. High Plains rise in elevation from around 1,800 to 7,000 ft, so the answer is 1,800 to 7,000 ft.\\
Answer &  1,800 to 7,000 ft\\
\\
Question &  Musician and satirist Allie Goertz wrote a song about the "The Simpsons" character Milhouse, who Matt Groening named after who?\\
Thought &  Milhouse was named after U.S. president Richard Nixon, so the answer is Richard Nixon.\\
Answer &  Richard Nixon\\
\\
Question &  Which documentary is about Finnish rock groups, Adam Clayton Powell or The Saimaa Gesture?\\
Thought &  Adam Clayton Powell (film) is a documentary about an African-American politician, not Finnish rock groups. So the documentary about Finnish rock groups must instead be The Saimaa Gesture.\\
Answer &  The Saimaa Gesture\\
\\
Question &  What profession does Nicholas Ray and Elia Kazan have in common?\\
Thought &  Professions of Nicholas Ray are director, screenwriter, and actor. Professions of Elia Kazan are director, producer, screenwriter, and actor. So profession Nicholas Ray and Elia Kazan have in common is director, screenwriter, and actor.\\
Answer &  director, screenwriter, actor\\
\\
Question &  Which magazine was started first Arthur's Magazine or First for Women?\\
Thought &  Arthur's Magazine was started in 1844. First for Women was started in 1989. 1844 (Arthur's Magazine) < 1989 (First for Women), so Arthur's Magazine was started first.\\
Answer &  Arthur's Magazine\\
\\
Question &  Were Pavel Urysohn and Leonid Levin known for the same type of work?\\
Thought &  Pavel Urysohn is a mathematician. Leonid Levin is a mathematician and computer scientist. So Pavel Urysohn and Leonid Levin have the same type of work.\\
Answer &  Yes\\

\midrule
\sethlcolor{orange}\hl{ReAct} &\\
& Solve a question answering task with interleaving Thought, Action, Observation steps. Thought can reason about the current situation, and Action can be two types: \\
&  (1) search[question], which searches a question on Google and returns a short snippet containing the answer. Note that sometimes the snippet does not contain the answer, and some alternative search might be needed. \\
&  (2) finish[answer], which returns the answer and finishes the task. \\
&  Here are some examples. \\
\\

Question &  What is the elevation range for the area that the eastern sector of the Colorado orogeny extends into?\\
\midrule \multicolumn{2}{r}{\textit{Continued on next page}} \\
\end{tabular}
\end{minipage}%
\end{table}
\begin{table}[h]
\scriptsize
\begin{minipage}{0.9\linewidth}
Hotpot QA Prompts -- Continued from previous page
\centering
\ttfamily
\begin{tabular}{l p{0.9\linewidth}}
\midrule
\\
Thought  &  I need to first find the eastern sector of the Colorado orogeny extends into what, then find its elevation range.\\
Action  &  search[the eastern sector of the Colorado orogeny extends into what?]\\
Observation 1 &  the High Plains\\
Thought  &  I need to find the elevation range for the High Plains.\\
Action  &  search[elevation range of the High Plains?]\\
Observation  &  around 1,800 to 7,000 ft\\
Thought  &  I have the answer.\\
Action  &  finish[1,800 to 7,000 ft]\\
Observation  &  Episode finished, reward = 1 \\
\\
Question &  Musician and satirist Allie Goertz wrote a song about the "The Simpsons" character Milhouse, who Matt Groening named after who?\\
Thought& I need to search Milhouse is named after who.\\
Observation& U.S. president Richard Nixon\\
Thought& I find the answer.\\
Action& finish[Richard Nixon]\\
Observation& Episode finished, reward = 1\\
\\
Question &   Which documentary is about Finnish rock groups, Adam Clayton Powell or The Saimaa Gesture? \\
Thought& I need to search documentary Adam Clayton Powell and documentary The Saimaa Gesture to find which is about Finnish rock groups.\\
Action& search[documentary Adam Clayton Powell]\\
Observation& Adam Clayton Powell (1989). Documentary. The Academy Award¨-nominated Adam Clayton Powell delves into the gripping life and career of the most influential ...\\
Thought& I do not get whether it is about Finnish rock groups. I need to search Adam Clayton Powell to make sure.\\
Action& search[Adam Clayton Powell]\\
Observation& Re-elected for nearly three decades, Powell became a powerful national politician of the Democratic Party, and served as a national spokesman on civil rights ...\\
Thought& Adam Clayton Powell is a politican, not Finnish rock groups. I need to search The Saimaa Gesture to make sure.\\
Action& search[The Saimaa Gesture documentary]\\
Observation& It is a documentary about three Finnish rock groups aboard the steamboat SS Heinävesi on their tour around Lake Saimaa. The Saimaa Gesture. Directed by, Aki ...\\
Thought& The Saimaa Gesture is about three Finnish rock groups, so the answer is The Saimaa Gesture.\\
Action& finish[The Saimaa Gesture]\\
Observation& Episode finished, reward = 1\\
\\

Question &  What profession does Nicholas Ray and Elia Kazan have in common?\\
Thought& I need to search the profession of Nicholas Ray and Elia Kazan, then find what is common.\\
Action& search[Nicholas Ray profession]\\
Observation& New York City, U.S.. Occupation(s), Film director, screenwriter, actor. Years active, 1946–1979. Spouses.\\
Thought& Nicholas Ray is film director, screenwriter, actor. I need to search Elia Kazan next.\\
Action& search[Elia Kazan profession]\\
Observation & Occupations. Actor; director; producer; screenwriter. Years active, 1934 - 1976. Spouses. Molly Day Thacher Kazan... (m. 1932, until her death in 1963).\\
Thought& Elia Kazan is actor, director, producer, screenwriter. So the common profession is actor, director, screenwriter\\
Action& finish[actor, director, screenwriter]\\
Observation& Episode finished, reward = 1\\

\midrule \multicolumn{2}{r}

\end{tabular}
\end{minipage}%
\end{table}
\begin{table}[ht]
\scriptsize
\begin{minipage}{0.9\linewidth}
MMLU Prompts
\centering
\ttfamily
\begin{tabular}{l p{0.9\linewidth}}
\midrule
\sethlcolor{cyan}\hl{Input-Output(IO)} & \\
Question &  Single choice :  A person takes buckets of water from the house and begins to add it to a pond in the yard. After a certain point, the pond\\
& A. bloats\\
& B. breaks\\
& C. sinks\\
& D. drowns\\
Answer &  A\\
\\
Question &  Single choice :  Coal is solid rock that began as organic material that was deposited in a swamp. The formation of coal suggests that, \\
& A. coal is made mostly of skeletal remains of animals.\\
& B. coal is formed from magma that has solidified over time.\\
& C. it quickly becomes petrified when water is removed.\\
& D. geologic processes continue over millions of years.\\
Answer &  D\\
\\
Question &  Single choice :  A student uses the following characteristics to describe a group of objects in space. \\
& * 200 billion stars \\
& * 30 million light years from Earth \\
& * 500 light years in diameter \\
& Which of the following is the student most likely describing?\\
& A. a galaxy\\
& B. the universe\\
& C. a constellation\\
& D. the solar system\\
Answer & A\\

\midrule
\\
\sethlcolor{yellow}\hl{Chain-of-Thoughts (CoT)} & \\
Question &  Single choice :  A person takes buckets of water from the house and begins to add it to a pond in the yard. After a certain point, the pond\\
& A. bloats\\
& B. breaks\\
& C. sinks\\
& D. drowns\\
Thought &  Each time the person adds a bucket of water, the level of water in the pond rises. Of all options, only A. bloats is consistent with the rise of water level. So the answer is A.\\
Answer &  A\\
Question &  Single choice:  Coal is solid rock that began as organic material that was deposited in a swamp. The formation of coal suggests that, \\
& A. coal is made mostly of skeletal remains of animals.\\
& B. coal is formed from magma that has solidified over time.\\
& C. it quickly becomes petrified when water is removed.\\
& D. geologic processes continue over millions of years.\\
Thought &  Let's evaluate each option. A. Coal is mostly composed of plant matter, not the skeletal remains of animals. Therefore, this option is incorrect. B. Coal is not formed from magma. Magma that solidifies over time creates igneous rocks, so this option is also incorrect. C. Petrification is a process by which organic material is turned into stone. It is not directly related to the process of coal formation, so this option is incorrect. D. The formation of coal takes incredibly long periods of time and consists of slow geologic processes such as sedimentation and metamorphism, which suggests that such processes continue over millions of years. Therefore, this option is correct. The answer is D.\\
Answer &  D\\
Question &  Single choice :  A student uses the following characteristics to describe a group of objects in space. \\
& * 200 billion stars \\
& * 30 million light years from Earth \\
& * 500 light years in diameter \\
& Which of the following is the student most likely describing?\\
& A. a galaxy\\
& B. the universe\\
& C. a constellation\\
& D. the solar system\\
Thought &  Let's evaluate each option. A. a galaxy :  Possibly, as galaxies do contain billions of stars and can be millions of light years from Earth. B. the universe :  Unlikely, as the universe is far larger than 30 million light years and contains more than just 200 billion stars. C. a constellation :  Unlikely, as constellations are patterns of stars seen from Earth and don't have a physical size or distance associated with them. D. the solar system :  Definitely not, as our solar system only contains one star, our sun. So, the answer is most likely A. a galaxy.\\
Answer & A\\
\midrule \multicolumn{2}{r}{\textit{Continued on next page}} \\
\end{tabular}
\end{minipage}%
\end{table}
\begin{table}[ht]
\scriptsize
\begin{minipage}{0.9\linewidth}
MMLU Prompts -- Continued from previous page
\centering
\ttfamily
\begin{tabular}{l p{0.9\linewidth}}
\midrule
\sethlcolor{orange}\hl{ReAct} &\\
& Solve a question answering task with interleaving Thought, Action, Observation steps. Thought can reason about the current situation, and Action can be two types:  \\
&  (1) search[question], which searches a question on Google and returns a short snippet containing the answer. Note that sometimes the snippet does not contain the answer, and some alternative search might be needed. \\
&  (2) finish[answer], which returns the answer and finishes the task. \\
&  Here are some examples. \\
\\
Question &  Single choice :  A person takes buckets of water from the house and begins to add it to a pond in the yard. After a certain point, the pond\\
& A. bloats\\
& B. breaks\\
& C. sinks\\
& D. drowns\\
Thought &  After continuously adding water to a pond, the pond will have more water than it could hold, thus bloats. So the answer is A.\\
Action &  finish[A]\\
Observation &  Episode finished, reward = 1 \\
\\
Question &  Single choice :  Coal is solid rock that began as organic material that was deposited in a swamp. The formation of coal suggests that, \\
& A. coal is made mostly of skeletal remains of animals.\\
& B. coal is formed from magma that has solidified over time.\\
& C. it quickly becomes petrified when water is removed.\\
& D. geologic processes continue over millions of years.\\
Thought &  The question is about the formation of coal. I need to first learn how coal is formed.\\
Action &  search[How is coal formed?]\\
Observation &  Coal takes millions of years to form Coal contains the energy stored by plants that lived hundreds of millions of years ago in swampy forests. Layers of dirt and rock covered the plants over millions of years. The resulting pressure and heat turned the plants into the substance we call coal.\\
Thought &  Based on the information, I can check each option. A :  coal is made by plants, not animals, so A is false. B :  I have no information about if coal is formed from magma yet. I could search "is coal formed from magma" to make sure. C :  I have no information about if coal quickly becomes petrified when water is removed. I could search "does coal quicklybecome petrified when water is removed" to make sure. D :  Coal takes millions of years to form, so D is possibly true. I could search "is the formulation of coal a geologic process" to make sure.\\
Action &  search[is the formulation of coal a geologic process]\\
Observation &  It is formed from plant remains that have been compacted, hardened, chemically altered, and metamorphosed by heat and pressure over geologic time.\\
Thought &  Seems the formulation of coal is over geologic time, so a geologic process. So the answer is D.\\
Action &  finish[D]\\
Observation &  Episode finished, reward = 1\\
\\
Question &  Single choice :  A student uses the following characteristics to describe a group of objects in space. \\
& * 200 billion stars \\
& * 30 million light years from Earth \\
& * 500 light years in diameter \\
& Which of the following is the student most likely describing?\\
& A. a galaxy\\
& B. the universe\\
& C. a constellation\\
& D. the solar system\\
Thought &  These options correspond to space systems of different sizes. I could search what is the diameter of each option to match.\\
Action &  search[what is the diameter of a galaxy]\\
Observation &  Most galaxies are 1,000 to 100,000 parsecs in diameter (approximately 3,000 to 300,000 light years) and are separated by distances on the order of millions of parsecs (or megaparsecs).\\

\midrule \multicolumn{2}{r}{\textit{Continued on next page}} \\

\end{tabular}
\end{minipage}%
\end{table}
\begin{table}[ht]
\scriptsize
\begin{minipage}{0.9\linewidth}
MMLU Prompts -- Continued from previous page
\centering
\ttfamily
\begin{tabular}{l p{0.9\linewidth}}
\midrule
Thought &  A galaxy is usually 3,000 to 300,000 light years in diameter, which is slightly more than 500 light years. I should search the diameter of the universe next.\\
Action &  search[what is the diameter of the universe]\\
Observation &  93 billion light-years\\
Thought &  The universe is 93 billion light years in diameter, which is much larger than 500 light years. I should search the diameter of a constellation next.\\
Action &  search[what is the diameter of a constellation]\\
Observation &  Its diameter, remarkably, is greater than 10 AU (1.5 billion kilometers!), large enough to fill the entire inner solar system almost as far out as Jupiter.\\
Thought &  A constellation is usually 10 AU in diameter. I need to convert it into light years.\\
Action &  search[10 AU to light years]\\
Observation &  0.000158125\\
Thought &  A constellation is usually 0.000158125 light years in diameter, which is much smaller than 500 light years. I should search the diameter of the solar system next.\\
Action &  Sedna is three times farther away from Earth than Pluto, making it the most distant observable object known in the solar system. It is 143.73 billion km from the Sun, thus giving the Solar System a diameter of 287.46 billion km.\\
Thought &  The solar system is 287.46 billion km in diameter. I need to convert it into light years.\\
Action &  search[287.46 billion km to light years]\\
Observation &  0.0303845459748716\\
Thought &  A constellation is usually 0.0303845459748716 light years in diameter, which is much smaller than 500 light years. Given all the information about diameters, the diameter of a galaxy is closest to 500 light years. So the answer is A.\\
Action &  finish[A]\\
Observation &  Episode finished, reward = 1\\
\midrule \multicolumn{2}{r}

\end{tabular}
\end{minipage}%
\end{table}
\begin{table}[ht]
\scriptsize
\begin{minipage}{0.9\linewidth}
Strategy QA Prompts
\centering
\ttfamily
\begin{tabular}{l p{0.9\linewidth}}
\midrule
\sethlcolor{cyan}\hl{Input-Output(IO)} & \\
Question &  Yes or no: Do the anchors on Rede Globo speak Chinese?\\
Answer &  no\\
\\
Question & Yes or no: Do the anchors on Rede Globo speak Chinese?\\
Answer &  no\\
\\
Question &  Yes or no: Would a student of the class of 2017 have amnesia about 9/11?\\
Answer &  yes\\
\\
Question & Yes or no: Is average number of peas in a pod enough commas for a billion?\\
Answer &  yes\\
\\
\midrule
\\
\sethlcolor{yellow}\hl{Chain-of-Thoughts (CoT)} & \\
Question & Yes or no: Do the anchors on Rede Globo speak Chinese?\\
Thought &The anchors on Rede Globo, a Brazilian television network, primarily speak Portuguese as that is the official language of Brazil. They may have proficiency in other languages, but Chinese is not likely to be one of the languages commonly spoken by the anchors on Rede Globo. So the answer is no.\\
Answer &no\\
\\
Question & Yes or no: Will the Albany in Georgia reach a hundred thousand occupants before the one in New York?\\
Thought &As of the most recent population estimates, Albany, New York, had a population of approximately 97,000 residents, while Albany, Georgia, had a population of around 73,000 residents. Albany, New York, is the capital of the state and is a major center for business, education, and government. It has a long history and economic significance, which attracts people to live and work in the area.On the other hand, Albany, Georgia, while an important regional center, is a smaller city in comparison. It does not have the same level of economic or cultural influence as Albany, New York. In conclusion, based on the current population figures and the different dynamics at play, it is unlikely that Albany, Georgia, will reach a population of one hundred thousand before Albany, New York. So the answer is no.\\
Answer &no\\
\\
Question & Yes or no: Is average number of peas in a pod enough commas for a billion?\\
Thought &Generally, a typical pea pod contains around 6 to 9 peas. A billion is a very large number 1,000,000,000 that requires 3 commas, which is less than the average number of peas in a pod. So the answer is yes.\\
Answer &yes\\
\\
Question & Yes or no: Is the language used in Saint Vincent and the Grenadines rooted in English?\\
Thought &Saint Vincent and the Grenadines were once British colonies, and English became the dominant language during the colonial period. After gaining independence in 1979, English remained as the official language of the country, and it has continued to be used in education, government, media, and daily communication. English has permeated various aspects of society and is widely spoken by the population, though local dialects and accents may influence the spoken form of English in the region. So the answer is yes. \\
Answer &yes\\
\\

\midrule
\sethlcolor{orange}\hl{ReAct} &\\
& Solve a question answering task with interleaving Thought, Action, Observation steps. Thought can reason about the current situation, and Action can be two types: \\
&  (1) search[question], which searches a question on Google and returns a short snippet containing the answer. Note that sometimes the snippet does not contain the answer, and some alternative search might be needed. \\
&  (2) finish[answer], which returns the answer and finishes the task. \\
&  Here are some examples. \\
\\
Question & Yes or no: Will the Albany in Georgia reach a hundred thousand occupants before the one in New York?
\\ Thought &I need to first find the population of Albany, Georgia, then find the population of Albany, New York, then compare them.
\\ Action& search[what is the current population of Albany, Georgia?]
\\ Observation & The current population of Albany, Georgia is 68,181 based on our projections of the latest US Census estimates.The last official US Census in 2020 recorded ...
\\ Thought &Albany, Georgia has 68,181 occupants in 2020.
\\ Action& search[what is the current population of Albany, New York?]\\

\midrule \multicolumn{2}{r}{\textit{Continued on next page}} \\
\end{tabular}
\end{minipage}%
\end{table}
\begin{table}[ht]
\scriptsize
\begin{minipage}{0.9\linewidth}
Strategy QA Prompts -- Continued from previous page
\centering
\ttfamily
\begin{tabular}{l p{0.9\linewidth}}
\midrule
\\
\\ Observation & The current population of Albany, New York is 97,593 based on our projections of the latest US Census estimates.The last official US Census in 2020 recorded ...
\\Thought &Albany, New York has 97,593 occupants in 2020, which is larger than Albany, Georgia. So Albany in Georgia will not reach a hundred thousand occupants before the one in New York, the answer is no.
\\ Action& finish[no]
\\ Observation & Episode finished, reward = 1

\\ Question & Yes or no: Do the anchors on Rede Globo speak Chinese?
\\ Thought &I need to know what is Rede Globo first.
\\ Action& search[what is Rede Globo?]
\\ Observation & TV Globo formerly known as Rede Globo, is a Brazilian free-to-air television network, launched by media proprietor Roberto Marinho on 26 April 1965.
\\ Thought &Rede Globo is a Brazilian television network, and Brazil is not a Chinese-speaking country, so anchors on Rede Globo do not speak Chinese.
\\ Action& finish[no]
\\ Observation & Episode finished, reward = 1

\\ Question & Yes or no: Would a student of the class of 2017 have amnesia about 9/11?
\\ Thought &The student's awareness about 9/11 would depend on their age at the time of the event, and if the age is too young, they would not have direct memory of the event. So, I need to first know how old is a student from class of 2017
\\ Action& search[when is a student from class of 2017 born?]
\\ Observation & The graduates of the class of 2017 were mostly born in 1999. Here's a look at the world they came up in. They are as old as The Phantom Menace. Midichlorians, Jar-Jar and pod racing have always been part of the Star Wars world for them.
\\ Thought &If a student is born around 1999, they would have been around 2 years old during the 9/11 attacks in 2001. I need to what age would have amnesia.
\\ Action& search[what age would have amnesia?]
\\ Observation & Although infants use their memories to learn new information, few adults can remember events in their lives that happened prior to the age of three. Psychologists at Emory University have now documented that age seven is when these earliest memories tend to fade into oblivion, a phenomenon known as "childhood amnesia."
\\ Thought &Amnesia happens for events prior to the age of three, so a student of the class of 2017 would have amnesia about 9/11.
\\ Action& finish[yes]
\\ Observation & Episode finished, reward = 1

\\ Question & Yes or no: Is average number of peas in a pod enough commas for a billion?
\\ Thought &I need to know the average number of peas in a pod, and the number of commas for a billion, then compare them.
\\ Action& search[what is the average number of peas in a pod?]
\\ Observation & Every pea seed grows into a single pea plant. An average pea plant will have 6 pods with 8 peas per pod, or 48 peas in total.
\\ Thought &The average number of peas in a pod is 8. I need to know how many commas in a billion.
\\ Action& search[how many commas in a billion?]
\\ Observation & A billion is expressed as '1,000,000,000', which amounts to three commas.
\\ Thought &The average number of peas in a pod (8 peas) is indeed greater than the number of commas used in a billion (3 commas), so the answer is yes.
\\ Action& finish[yes]
\\ Observation & Episode finished, reward = 1.
\\
\midrule \multicolumn{2}{r}

\end{tabular}
\end{minipage}%
\end{table}

\end{document}